\documentclass[11pt,a4paper]{article}
\usepackage[hyperref]{emnlp2020}
\usepackage{times}
\usepackage{latexsym}

\usepackage{float}

\usepackage{graphicx}
\usepackage{booktabs}
\usepackage{eqparbox}
\usepackage{arydshln}
\usepackage{multirow}
\usepackage{amsmath}
\usepackage{enumitem}
\usepackage{array}

\usepackage{microtype}

\aclfinalcopy 

\newcommand*\samethanks[1][\value{footnote}]{\footnotemark[#1]}

\title{Knowledge Guided Named Entity Recognition for BioMedical Text}

\author{ Pratyay Banerjee \thanks{\quad These authors contributed equally to this work.}  \: Kuntal Kumar Pal\samethanks \:\: Murthy Devarakonda \: Chitta Baral 
\\ Department of Computer Science, Arizona State University
\\ \texttt{pbanerj6,kkpal,mdevarak,chitta}@asu.edu
}

\date{}

\begin{document}
\maketitle
\begin{abstract}
In this work, we formulate the NER task as a multi-answer knowledge guided QA task (KGQA) which helps to predict entities  only by assigning B, I and O tags without associating entity types with the tags. We provide different knowledge contexts, such as, entity types, questions, definitions and examples along with the text and train on a combined dataset of 18 biomedical corpora. 
This formulation (a) enables systems to jointly learn NER specific features from varied NER datasets, (b) can use knowledge-text attention to identify words having higher similarity to provided knowledge, improving performance, (c) reduces system  confusion by reducing the prediction classes to B, I, O only, and
(d) makes detection of nested entities easier. We perform extensive experiments of this KGQA formulation on 18 biomedical NER datasets, and through experiments we note that knowledge helps in achieving better performance. 
Our problem formulation is able to achieve state-of-the-art results in 12 datasets.
\end{abstract}

\section{Introduction}

\begin{figure}[ht]
  \includegraphics[width=\linewidth]{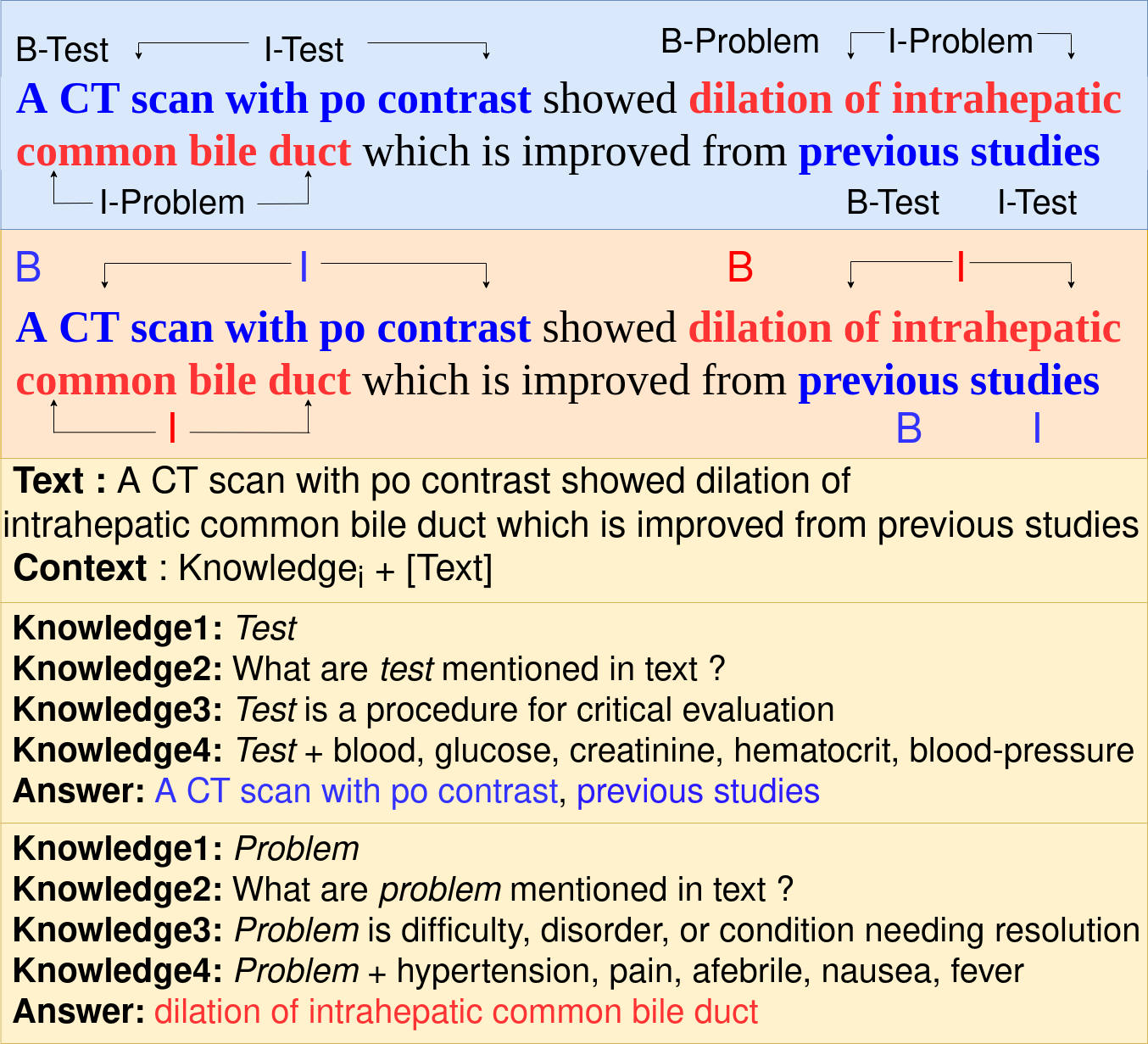}
  \caption{The top block shows the traditional way of NER. In our method, we predict only B, I and O tags for a given context, i.e, only red tags are predicted if the context is about entity Problem. }
  \label{fig:appr}
\end{figure}

Named Entity Recognition (NER) has been considered as a relatively difficult task in biomedical domain due to the stylized writing and domain-specific terminology. Moreover, the target entities are usually proper nouns or unregistered words, with new words for drugs, diseases, and chemicals being generated frequently. Also, the same word phrases can be recognized as different named entities in terms of current context \cite{cohen2004natural, liu2006quantitative,song2018comparison}.  For these reasons, external knowledge can be helpful to guide automated systems to identify the entities in biomedical domain.

In general domain, use of extensive external knowledge has helped systems in multiple natural language tasks like commonsense question answering \cite{sap2019socialIQa,talmor-etal-2019-commonsenseqa} and science question answering \cite{lai2017race,OpenBookQA2018}. In biomedical named entity recognition, external knowledge can be about entities and their relations. Knowledge like entity-types along with their definition and examples can allow attention mechanism to compare and learn to detect new entities, especially if the entity is newly generated and has infrequent mentions in common biomedical texts. Apart from the use of external knowledge, framing an NLP task as a question answering task can lead to better performance \cite{mccann2018natural}. Motivated by this approach, we hypothesize that a Knowledge guided QA framework may be helpful in biomedical NER.

In this paper, we focus on NER in biomedical text and we test our hypothesis using different kinds of knowledge. The ways of expressing knowledge include asking a question about the entity, giving the entity type, providing a definition of the entity type and mentioning some examples of the entity types, as seen in Figure  \ref{fig:appr}. 

Figure \ref{fig:appr} also shows, how traditional NER systems formulate the problem as a classification task. This traditional task formulation leads to the following challenges: (a) \textit{labelling error}, i.e., even though a system is able to identify the location of an entity correctly, it fails to predict the correct type; (b) inability to leverage more information for a particular entity type, since the conventional task formulation only allows to predict all entity types jointly; (c) lack of labelled data for each entity type, especially in the biomedical domain. Challenge (a) and (b) are even more profound in the presence of nested named entities.

We can avoid challenges (a) and (c) by modeling the task as multi-answer extraction task, where we predict only one type of entity at a time, given a context determining which entity is being extracted at the current time. This formulation allows us to avert the issue of nested named entities and  helps us to jointly learn from multiple biomedical datasets having similar entities. We specifically address the challenge (b) by providing four different types of knowledge context as shown in Figure \ref{fig:appr}.
We perform an empirical study of which knowledge type has the most significant impact in a NER task. 

Our task formulation enables us to create a considerably large dataset with knowledge context utilizing 18 biomedical datasets. The goal is to learn jointly from multiple domains containing different target entities. We also propose a new NER model over BERT using a re-contextualization layer called BERT-CNN. This layer uses token-local features to recompute token encoding to enable the model to better understand the start and end locations of an entity. We use the BERT-base model and show our task formulation and model performs better than a strong baseline of a BERT-large model pre-trained on medical corpus, and finetuned using the traditional NER task. 
We perform extensive experiments to analyze the impact of each of our contributions. We also study the transfer learning ability of our knowledge guided BERT-CNN model, as one of the major challenges currently faced by the biomedical community is the poor ability of the models to transfer in real-life applications. 

To summarize our contributions: 

\begin{itemize}[noitemsep] 
    \item We reformulate the task of named entity recognition as a multi-answer question answering task using knowledge as a context.  
    \item We make available a significantly large, cleaned and pre-processed dataset with knowledge context utilizing 18 biomedical datasets having in total 398495 training, 148166 validation and 502306 test samples.  
    \item We propose a BIO tagging based model for the knowledge guided named entity recognition task, with a re-contextualization layer. 
    \item We perform extensive experiments to evaluate our models, including the ability of the model to adapt to new domains. 
    \item Finally, all our contributions together further push the state-of-the-art exact match F1 scores by 1.5-11\% for 12 publicly available biomedical NER datasets. 
\end{itemize}

\section{Our Approach }
\subsection{Task Formulation}
Traditional systems define named entity recognition as a multi-class classification task. Given a context $C=\{c_1, c_2, ..., c_n\}$, any token $c_i$ is classified as one of the three tags $B$-$e_k$, $I$-$e_k$, $O$ in the BIO-Tagging scheme, where $e_k$ $\in$ $E$ (the set of  entity types for a dataset). This formulation leads to \textit{labelling error}. A token $c_i$ is classified as $B$-$e_k$ or $I$-$e_k$ when the token is actually a $B$-$e_j$ or $I$-$e_j$ where $j\neq k$. This means that even though a system was able to identify the location of an entity correctly, it fails to identify the correct type. 

In our approach, we tackle this issue by formulating the NER task in the following way. Given a context $C=\{c_1, c_2, ..., c_n\}$, any token $c_i$ is classified as $B$, $I$ and $O$. To identify which entity, type the token belongs to, we provide external knowledge $K$ to the context which in turn contains the entity type information. For example, if we want to extract two entities $e_1$ and $e_2$ from context $C$, we first provide $K_{e_1}$ and $C$ as input to our model to extract $e_1$ entities, then provide $K_{e_2}$ and $C$ as input to extract $e_2$ entities. This formulation decouples the classification and the entity location tasks, enabling the model to learn from multiple datasets and overcoming \textit{labelling error}. 

\subsection{Knowledge Context Generation}
We experiment with five types of knowledge context ($K$)  to identify entities and their types. These are: (a) \textit{Entity types} ($e_k \in E$)  (b) separate \textit{Question} ($Q_k$) created using each entity type,  (c) \textit{Definition} ($D_k$) of each entity type along with the entity type itself, (d) \textit{Examples} ($Eg_k$) along with entity type and (e) \textit{All} of the above. If there are entities of $n$ entity types in a text, during training we create a set of five knowledge context for each different entity type. Since the approach works one entity type at a time, we make sure that the entity type is mentioned in each of the five contexts. During inference, only the best knowledge context is used, i.e, if \textit{Question} performs best for a dataset, we use only that context.   
For the example mentioned in Figure \ref{fig:appr},  
$E$ is \{``\textit{Problem}", ``\textit{Test}"\} , $Q$ is \{``\textit{What are problem mentioned in text?}", ``\textit{What are test mentioned in text?}"\}, $D$ is the definition text, \{``\textit{Problem is a difficulty, disorder, or condition needing resolution}", ``\textit{Test is a procedure for critical evaluation}''\}, and $Eg$ are the examples \{``\textit{hypertension, pain, afebrile, nausea, fever}", ``\textit{blood, glucose, creatinine, hematocrit, blood-pressure}''\}. 


\subsection{Datasets}
We create the dataset for NER using fifteen publicly available biomedical datasets\footnote{https://github.com/cambridgeltl/MTL-Bioinformatics-2016}\cite{crichton2017neural} and three datasets from previous i2b2  challenges \cite{sun2013evaluating,uzuner20112010,uzuner2010extracting,uzuner2012evaluating}. 
One of the samples is shown in Figure \ref{fig:appr}. Our task formulation enables us to combine the datasets and a create a significantly large dataset, that enables deep neural model learning. Moreover, the multi-task learning for different entity types enables the model to generalize better. 

\noindent
\textbf{Bionlp Shared Task and Workshop}: Six of the datasets Bionlp09, Bionlp11ID, Bionlp11EPI, Bionlp13PC, Bionlp13CG, Bionlp13GE are from the Biomedical Natural Language Processing Workshops.
Some of the basic entities of these datasets  are gene  or gene products, protein, chemicals and organisms.

\noindent
\textbf{i2b2 Shared Task and Workshop}: We use three datasets from i2b2 shared task and Workshop Challenges in Natural Language Processing for Clinical Data. We only use training and testing data from 2010 Relations Challenge, 2011 Coreference Challenge and 2012 Temporal Relations Challenge. These datasets primarily contain entities like problems, tests and treatments.

\noindent
\textbf{Bio-Creative Challenge and Workshop}: These  workshops  provide datasets for information extraction task in biological domain. We only use three datasets namely BC4CHEMD (Chemical), BC5CDR (Chemical and disease) and BC2GM (gene or protein). We consider these datasets since they are similar to biomedical texts and can be augmented to be trained together to generalize on extraction of some of the entities.

\noindent
\textbf{Others}: Apart from these 12 datasets we also include CRAFT, AnatEM, Linnaeus, JNLPBA, Ex-PTM and NCBI-Disease to increase our training and evaluation set. They include entities such as anatomy, species, diseases, cell-line, DNA, RNA, gene or protein and chemicals.

\subsection{Rule-based Template Creation}
We use the following rules to create contexts for each knowledge type.

\noindent
\textbf{Entity: }The first and the simplest context, is the Entity type name itself.

\noindent
\textbf{Question:} We create a knowledge context Question ($Q_k$), using simple rules, like:

$Q_k$ = What are the $[e_k]$ mentioned in the text ?

\noindent
\textbf{Definition:} To get knowledge context ($D_k$), we find the corresponding scientific definition of each entity type from UMLS Meta-thesaurus by considering the entity type as concept \cite{bodenreider2004unified}. Other sources include challenge dataset definitions and online resources.

$D_k$ $\in$ $\{$UMLS $\vert$ Challenge $\vert$ Online Resource$\}$

\noindent
\textbf{Examples:} 
In order to determine representative examples of an entity type, we find the top ten most frequent entities for each type from the entire training dataset. We concatenate these ten entities as the final knowledge context ($Eg_k$) and prepend this in front of the text.

\noindent
\textbf{All:} is just the concatenation of all of the above knowledge context.


The motivation behind using the entity type and definition as knowledge context is that the neural model can leverage the information present in the knowledge to attend to correct entities. If we use question as a context then the task becomes a multi-answer question-answering task. We use examples as knowledge with the hypothesis that our model will be able to choose the entities that can belong to same categories as the examples. 

The distribution of each of the entities across each of the dataset for Training, Validation and Test splits (both positive and negative samples) and more details about the dataset preparation can be found in the Supplemental Material. 

We treat each individual sentence in a medical document or paragraph as an individual sample. If a sentence has an entity corresponding to a context, we consider that as a positive sample for that context. Similarly, we treat a sentence that does not have an entity for a corresponding context as a negative sample for that context. Although these sentences can contain entities for other entity types. Since many datasets do not provide a validation split, we randomly sample from the train split to create our validation data. Overall, our dataset has 398495 train, 148166 dev and 502306 test samples.  



\begin{figure}[ht]
  \includegraphics[width=\columnwidth]{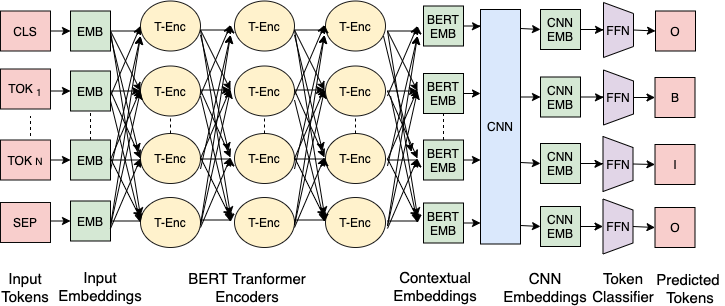}
  \caption{ BERT-CNN for Multi-Answer KGQA}
  \label{fig:corr}
\end{figure}

\begin{table*}[t]
\centering
\tiny
\resizebox{\textwidth}{!} {
\begin{tabular}{llccccccccc}
\toprule
\multicolumn{2}{c}{} & \textbf{ANATEM}     & \textbf{BC2GM} & \textbf{BC4CHEMD} & \textbf{BC5CDR} & \textbf{BIONLP09} & \textbf{BIONLP11EPI}   \\ \hline
\multirow{3}{*}{\rotatebox[origin=c]{90}{\textbf{BASE}}}  & \textbf{BioBERT}    &   
89.63 \;		89.32 \;		89.47         &  	
82.45 \;		83.83 \;		83.13    &     	
90.97 \;	89.59 \;	90.27\;     &        
87.19 \;		90.59 \;		88.84&  
89.65 \;		88.60 \;		89.13&
85.23 \;		85.60 \;		85.41
\\ 
                           & \textbf{MimicBERT}   &   
86.23 \;		85.72 \;		85.98         &  	
79.04 \;	80.32 \;		79.68     &     	
88.77 \;	85.41 \;		87.06     &        
84.01 \;	86.86 \;		85.39&          
87.71 \;	84.15 \;		85.89&  
79.37 \;	77.78 \;	78.57   \\ 
                           & \textbf{BERT-MRC}   &
72.24 \;	75.09 \;	73.64&
73.80 \;	74.59 \;	74.19&
86.47 \;	85.52 \;	85.99&
71.22 \;	73.68 \;	72.43  &       
74.62 \;	70.69 \;	72.60&         
77.81 \; 67.01 \; 72.01\\ \hline
\multicolumn{2}{c}{\textbf{SOTA}}             &   
\;\;\;\;\;- \;\;\;\;\;\;	- \;\;\;\;\;	\textbf{91.61}*  &  
\;\;\;\;\;- \;\;\;\;\;\;	- \;\;\;\;\;	81.69*       &      	
\;\;\;\;\;- \;\;\;\;\;\;	- \;\;\;\;\;		\textbf{92.36}     &    	
\;\;\;\;\;- \;\;\;\;\;\;	- \;\;\;\;\;		\textbf{90.01}     &     
\;\;\;\;\;- \;\;\;\;\;\;	- \;\;\;\;\;		84.20*&          
\;\;\;\;\;- \;\;\;\;\;\;	- \;\;\;\;\;		78.86* \\ \hline
\multirow{3}{*}{\rotatebox[origin=c]{90}{\textbf{OURS}}}                       & \textbf{BioBERT}    &  
90.29 \;	89.43 \;	\underline{89.85} &   
82.47 \;	83.36 \;	82.91 &     
91.93 \;	91.11 \;	91.52     &   
89.63 \;	88.80 \;	89.21&       
91.35 \;	92.21 \;	91.78 &     
88.26 \;    86.77 \;    87.51       \\ 
                           & \textbf{MimicBERT}  &    
87.05 \;	86.50 \;	86.80&
81.22 \;	81.40 \;	81.31&
89.47 \;	88.86 \;	89.16&
88.25 \;	86.78 \;	87.51&
89.19 \;	91.41 \;	90.29&
88.01 \;	82.19 \;	85.00\\ 
                           & \textbf{BERT-CNN}   & 
89.78 \;  89.24  \;  89.51& 
82.89 \;   83.39 \;   \textbf{83.14}$\dagger$&
92.56 \;	91.10 \;	\underline{91.82}&
90.09 \;   89.16 \; \underline{89.62}&
91.55 \;   92.95 \; \textbf{92.25}$\dagger$ &
88.58 \;   87.40 \; \textbf{87.99}$\dagger$ \\ 
\toprule
\multicolumn{2}{c}{} & \textbf{BIONLP11ID} & \textbf{BIONLP13CG} & \textbf{BIONLP13GE} & \textbf{BIONLP13PC} & \textbf{CRAFT} & \textbf{EXPTM}    
\\ \hline
\multirow{3}{*}{\rotatebox[origin=c]{90}{\textbf{BASE}}}  & \textbf{BioBERT}     &   
84.23 \;	85.77 \;	84.70         &  	
84.82 \;	86.42 \;		85.56     &     	
72.92 \;	85.42 \;		78.68     &        
87.64 \;	90.56 \;		89.06&          
84.92 \;	86.56 \;		85.70&  
76.12 \;	79.81 \;	77.92   \\ 
                           & \textbf{MimicBERT}  &   
83.93 \;	81.84 \;	82.35         &  	
77.52 \;	80.32 \;	78.71     &     	
64.72 \;	65.53 \;		65.12     &        
81.59 \;	85.45 \;	83.43&          
81.27 \;	79.08 \;		80.04&  
66.74 \;	67.29 \;	67.01
\\   
                           & \textbf{BERT-MRC}   &
80.25 \;	73.26 \;	76.60 &
74.57 \;		69.25 \;		71.81&
77.79 \;	75.54 \;	76.65&
76.51 \;	76.95 \;	76.73 &
75.79 \;	72.25 \;	73.98&
76.61 \;	76.93 \;	76.77\\ \hline
\multicolumn{2}{c}{\textbf{SOTA}}                &   
\;\;\;\;\;- \;\;\;\;\;\;	- \;\;\;\;\;	81.73*  &  
\;\;\;\;\;- \;\;\;\;\;\;	- \;\;\;\;\;	78.90*       &      	
\;\;\;\;\;- \;\;\;\;\;\;	- \;\;\;\;\;		78.58*     &    	
\;\;\;\;\;- \;\;\;\;\;\;	- \;\;\;\;\;		81.92*     &     
\;\;\;\;\;- \;\;\;\;\;\;	- \;\;\;\;\;		79.56*&          
\;\;\;\;\;- \;\;\;\;\;\;	- \;\;\;\;\;		74.90* 
\\ \hline
\multirow{3}{*}{\rotatebox[origin=c]{90}{\textbf{OURS}}}      & \textbf{BioBERT}    &    
86.34 \;	85.58 \;	85.96&     
87.18 \;	87.28 \;	87.23&         
82.28 \;	86.58 \;	\underline{84.38} &     
90.14 \;	92.09 \;	\textbf{91.11}$\dagger$    &       
88.18 \;	88.61 \;	88.39&      
85.97 \;	85.30 \;	\textbf{85.64}$\dagger$       \\ 
                           & \textbf{MimicBERT}  &     
83.12 \;	81.78 \;	82.45& 
85.08 \;	85.37 \;	85.23&   
81.61 \;	86.28 \;	83.88&    
87.62 \;	89.63 \;	88.61&   
85.01 \;	87.14 \;	86.06&   
84.09 \;	81.34 \;	82.69\\
                           & \textbf{BERT-CNN}   & 
87.98 \;	84.64 \;	\textbf{86.27}$\dagger$ &
90.62 \;   88.56 \;   \textbf{89.58}$\dagger$ &
83.77 \;   88.01 \;   \textbf{85.84}$\dagger$ &
89.03 \;   91.87 \;   90.43&
90.54 \;   89.19 \;   \textbf{89.86}$\dagger$ & 
85.08 \;   84.79 \;   84.94\\ 
\toprule
\multicolumn{2}{c}{}  & \textbf{JNLPBA} & \textbf{LINNAEUS} & \textbf{NCBIDISEASE} & \textbf{2010-i2b2}  & \textbf{2011-i2b2}  & \textbf{2012-i2b2}  \\ \hline
\multirow{3}{*}{\rotatebox[origin=c]{90}{\textbf{BASE}}}  & \textbf{BioBERT}      &   
69.96 \;   	78.19 \;   	73.63         &  	
92.30 \;   	86.42 \;   	89.27         &  	    	
86.67 \;   	89.38 \;   	88.00         &  
85.32 \;	83.23 \;	84.26&
91.24 \;	90.32 \;	90.78& 
79.31 \;	75.89 \;	77.56           \\ 
                           & \textbf{MimicBERT}    &   
67.99 \;	76.32 \;	71.66         &  	
91.69 \;	81.81 \;	86.46         &  	    	
84.04 \;	88.23 \;	86.08         &  	     
90.37 \;	88.29 \;	89.32         &  	       
92.83 \;	91.22 \;	92.02         &  
79.78 \;	81.01 \;	80.39            \\ 
                           & \textbf{BERT-MRC}   & 
70.52 \;	69.38 \;	69.95&
74.13 \;	73.56 \;	73.84&
77.25 \;	73.23 \;	75.19&
75.32 \;	73.23 \;	74.26&
81.24 \;	80.32 \;	80.78& 
69.31 \;	65.89 \;	67.56\\ \hline
\multicolumn{2}{c}{\textbf{SOTA}}                  &   
\;\;\;\;\;- \;\;\;\;\;\;	- \;\;\;\;\;	78.58*  &  
\;\;\;\;\;- \;\;\;\;\;\;	- \;\;\;\;\;	\textbf{95.68}*       &      	
\;\;\;\;\;- \;\;\;\;\;\;	- \;\;\;\;\;		89.36     &    	
\;\;\;\;\;- \;\;\;\;\;\;	- \;\;\;\;\;		90.25\#     &     
\;\;\;\;\;- \;\;\;\;\;\;	- \;\;\;\;\;		-&          
\;\;\;\;\;- \;\;\;\;\;\;	- \;\;\;\;\;		80.91\# 
\\ \hline
\multirow{3}{*}{\rotatebox[origin=c]{90}{\textbf{OURS}}}      & \textbf{BioBERT}    & 
76.12 \;	82.15 \;	79.02&         
90.32 \;	89.88 \;	90.10&  
87.50 \;	90.67 \;	89.05&   
93.29 \;	94.41 \;	93.84&      
93.88 \;	94.15 \;	94.02& 
73.53 \;	83.21 \;	78.07\\ 
                           & \textbf{MimicBERT}  &    
74.97 \;	80.79 \;	77.77&   
86.31 \;	85.10 \;	85.70&     
86.82 \;	88.80 \;	87.80&    
94.85 \;	95.76 \;	95.30&        
94.18 \;	94.30 \; 94.24&
81.57 \;	84.76 \;	83.13\\ 
                           & \textbf{BERT-CNN}   & 
76.85 \;	81.79 \;	\textbf{79.24} &
90.69 \;   90.53 \;   \underline{90.61}&
87.89 \;   91.56 \;   \textbf{89.69}&
95.27 \;   95.91 \;   \textbf{95.59}$\dagger$ &
94.70 \;   94.94 \;   \textbf{94.82}&
84.83 \;   85.25 \;   \textbf{85.04}$\dagger$ 
\\ \hline
\end{tabular}
}
\caption{Precision, Recall and F-Measure (in order) for 18 datasets compared with multiple models. * tagged scores are non-BERT systems, \#  BERT-Large and rest are BERT-Base systems. Best F1-scores are in bold. Underlined are our best scores where our models are not SOTA. $\dagger$ tagged scores are statistically significantly better than SOTA ($p\leq0.05$  based on Wilson score intervals \cite{wilson1927probable}). For dataset statistics please see the Supp. Materials.}
\label{tab:sota}
\end{table*}

\section{Model Description}

\subsection{Knowledge Guided NER}
We choose the BERT-base cased version \cite{devlin2018bert} as our base model. In our approach, given a text $C$, we create a knowledge context $K_i$ for each context type. We need to find the spans of entities $S_{start}$ and $S_{end}$.  
So, we define the input to the BERT model as follows, the knowledge context tokens $K_i=\{k_{ij}\}$ are prepended to the text tokens, $C=\{c_j\}$. 
The sequence of tokens, $\{[CLS],k_{i1},..k_{im},[SEP],c_1,..c_n,[SEP]\}$ is given as input to the BERT model where $m$ is the size of knowledge context $K_i$ and $n$ is the size of text $C$. 
In our baseline model, for each token we predict $B$, $I$ and $O$ using a feed-forward layer. 

\subsection{Re-contextualization}
We modify the BERT-base model by adding a re-contextualization layer consisting of a two-dimensional convolution layer. The purpose of this layer is to leverage information from adjacent or token-local embeddings and help in better start and end prediction of the entities. As BERT uses multiple layers of attention which jointly focuses on all the tokens, we add this CNN layers with a window of $W<5$ to focus on nearby tokens only. We take the outputs of the CNN layer and feed it to the final feed-forward layer to predict the tags.
Figure \ref{fig:corr} represents the end-to-end architecture of our BERT-CNN model.
 
\subsection{Training and Testing}
During training, the context, $X$ (combination of knowledge, $K_i$  and given text, $C$) has gold annotations ($y_i$) of $B$, $I$ and $O$ for each token ($x_i)$. We calculate cross-entropy loss for each token $x_i$ as:
\begin{equation*}
   L_{token} = - \sum_{c=1}^{M} y_{x_i,c} log(P_{{x_i},c})
\end{equation*}
where $M$ is the total number of classes (B, I, O), $y_{x_i,c}$ is a binary indicator whether the label $c$ is the correct classification of token $x_i$, $P_{{x_i},c}$ is the predicted probability of $x_i$ belonging to class $c$.

The model is trained end-to-end with the above loss. During inference we consider the tokens $x_i$ present only in the text $C$. Text chunks that start from label $B$ and continue till last $I$ tag are predicted as entities. For each entity type we feed the text with a separate context and text input. We train our model jointly on a processed combined dataset of 18 common biomedical datasets and compare the performance when trained individually.

\section{Experiments}

\begin{table*}[t]
\tiny
\centering
\resizebox{\textwidth} {!}{
\begin{tabular}{lccccc|ccccc|ccccc}
\toprule

\multirow{2}{*}{\textbf{DATASET}} & \multicolumn{5}{c}{\textbf{PRECISION}} & \multicolumn{5}{c}{\textbf{RECALL}} & \multicolumn{5}{c}{\textbf{F-MEASURE}} \\ \cline{2-16} 
                         & \textbf{T}       & \textbf{Q}       & \textbf{D}       & \textbf{E} &\textbf{A}       
                         & \textbf{T}       & \textbf{Q}       & \textbf{D}       & \textbf{E}      &\textbf{A}  
                         & \textbf{T}       & \textbf{Q}       & \textbf{D}       & \textbf{E}       &\textbf{A}             \\ \hline

\shortstack{\textbf{\texttt{ANATEM}}} &\underline{89.43} & 88.81 & 89.03 & 89.41 & \textbf{89.78} & 87.54 & \underline{89.12} & 87.23 & 88.69 & \textbf{89.24} & 88.47 & 88.96 & 88.13 & \underline{89.05} & \textbf{89.51} \\
\shortstack{\textbf{\texttt{BC2GM}}} &81.79 & \textbf{82.89} & 80.96 & \underline{82.37} & 81.17 & \underline{82.32} & \textbf{83.39} & 81.45 & 82.29 & 82.06 & 82.05 & \textbf{83.14} & 81.21 & \underline{82.33} & 81.61 \\ 
\shortstack{\textbf{\texttt{BC4CHEMD }}}&90.25 & \textbf{92.07} & 89.81 & 90.57 & \underline{91.19} & 88.48 & \textbf{91.01} & 88.13 & 89.27 & \underline{90.49} & 89.36 & \textbf{91.54} & 88.96 & 89.92 & \underline{90.84} \\ 
\shortstack{\textbf{\texttt{BC5CDR}}} &87.93 & \textbf{90.09} & 88.07 & 88.49 & \underline{89.01} & 86.72 & \textbf{89.16} & 86.12 & 86.62 & \underline{87.83} & 87.32 & \textbf{89.62} & 87.08 & 87.55 & \underline{88.42} \\
\shortstack{\textbf{\texttt{BIONLP09}}}&88.86 & \underline{90.85} & 51.14 & 89.75 & \textbf{91.55} & 89.35 & \underline{92.75} & 69.14 & 89.05 & \textbf{92.95} & 89.11 & \underline{91.78} & 58.79 & 89.40 & \textbf{92.25} \\
\shortstack{\textbf{\texttt{BIONLP11EPI}}}&86.49 &\textbf{ 88.58} & 77.56 & 87.55 & \underline{87.94} & 83.74 & \textbf{87.40} & 83.53 & 83.32 & \underline{85.75} & 85.09 & \textbf{87.99} & 80.44 & 85.38 & \underline{86.83} \\
\shortstack{\textbf{\texttt{BIONLP11ID}}}&\underline{86.19} & \textbf{86.60} & 84.29 & 85.14 & 85.56 & 81.09 & \textbf{85.35} & 81.24 & 81.05 & \underline{83.38} & 83.56 & \textbf{85.97} & 82.74 & 83.04 & \underline{84.46} \\
\shortstack{\textbf{\texttt{BIONLP13CG}}}&88.21 & \underline{89.49} & 87.45 & 89.04 & \textbf{90.62} & 83.23 & \underline{86.45} & 82.71 & 85.57 & \textbf{88.56} & 85.65 & \underline{87.94} & 85.01 & 87.27 & \textbf{89.58} \\
\shortstack{\textbf{\texttt{BIONLP13GE}}} &80.86 & \textbf{83.77} & 68.64 & 80.26 & \underline{83.25} & 84.72 & \textbf{88.01} & 81.32 & 84.48 & \underline{85.36} & 82.74 & \textbf{85.84} & 74.44 & 82.32 & \underline{84.29} \\
\shortstack{\textbf{\texttt{BIONLP13PC}}}&\underline{89.79} & 89.03 & 88.43 & 88.95 & \textbf{89.87} & 89.55 & \textbf{91.87} & 88.33 & 90.58 & \underline{90.60} & 89.67 &\textbf{ 90.43} & 88.38 & 89.76 & \underline{90.23} \\
\shortstack{\textbf{\texttt{CRAFT}}} &86.81 & \underline{88.07} & 82.79 & 88.00 & \textbf{90.54} & 84.31 & 88.19 & 84.67 & \underline{89.18} & \textbf{89.19} & 85.54 & 88.13 & 83.72 & \underline{88.58} & \textbf{89.86} \\
\shortstack{\textbf{\texttt{EXPTM}}}&\underline{84.27} & 84.26 & 74.01 & 84.06 & \textbf{85.08} & 82.67 &\textbf{ 85.39} & 83.96 & 82.72 & \underline{84.79} & 83.65 & \underline{84.83} & 78.67 & 83.38 & \textbf{84.94} \\
\shortstack{\textbf{\texttt{JNLPBA}}}&71.56 & \textbf{76.64} & 68.07 & 71.68 & \underline{75.79} & 77.63 & \textbf{80.97} & 70.41 & 78.42 & \underline{80.36} & 74.48 & \textbf{78.75} & 69.22 & 74.89 & \underline{78.01} \\
\shortstack{\textbf{\texttt{LINNAEUS}}} &\underline{91.34} & 88.47 & 86.33 & \textbf{92.37} & 90.69 & 86.01 & \underline{88.47} & 86.52 & 88.30 & \textbf{90.53} & 88.59 & 88.47 & 86.43 & \underline{90.29} & \textbf{90.61} \\
\shortstack{\textbf{\texttt{NCBIDISEASE}}} &86.64 & \textbf{87.94} & 86.99 & 86.33 & \underline{87.89} & \underline{90.05} & 89.94 & 89.43 & 89.84 & \textbf{91.56} & 88.31 & \underline{88.93} & 88.19 & 88.05 & \textbf{89.69} \\
\shortstack{\textbf{\texttt{2010-i2b2}}}&93.42 & \textbf{95.27} & 93.06 & 93.43 & \underline{94.87} & 94.13 & \textbf{95.91} & 93.88 & 94.60 & \underline{95.66} & 93.77 & \textbf{95.59} & 93.47 & 94.01 & \underline{95.26} \\
\shortstack{\textbf{\texttt{2011-i2b2}}}  &93.10 & \underline{94.42} & 92.66 & 94.23 & \textbf{94.70} & 92.04 & \underline{94.37} & 92.05 & 93.67 & \textbf{94.94} & 92.57 & \underline{94.40} & 92.35 & 93.95 & \textbf{94.82} \\
\shortstack{\textbf{\texttt{2012-i2b2}}}   &\underline{82.27} & \textbf{84.83} & 81.00 & 75.63 & 67.23 & 81.27 & \textbf{85.25} & 81.33 & 83.96 & \underline{84.00} & \underline{81.77} & \textbf{85.04} & 81.17 & 79.58 & 74.68 \\
\hline
\end{tabular}
}
\caption{Precision, Recall and F-Measure of BERT-CNN model using different knowledge types: Entity Type (T), Question (Q), Definition (D), Examples (E) and all of them together (A). Best scores are in bold, second best are underlined. Mean of three random seed runs are reported.
 }
\label{tab:knowledge}
\end{table*}

\subsection{Experimental Setup and Training Parameters}
We use a batch size of 32 and a learning rate of 5e-5 for all our experiments. The maximum sequence length of 128/256 depends on the 99th percentile of the input token lengths. We train using 4 NVIDIA V100 16GB GPUs, with a patience of 5 epochs. We report the mean F1 scores for three random seeds, the deviation is reported in Supplemental Materials. For BERT-CNN model, we apply a two-dimensional convolution layer on top of BERT contextual token embeddings. The convolution layer uses a $5\times5$ size kernel. The stride size is (1,2), where 1 is across sentence dimension, and 2 is across word embedding dimension. We also perform circular padding. 
        
\subsection{Baseline Models}
We consider the following models as strong baselines for our work. The first set of baselines are the BERT models pre-trained on biomedical text BioBERT and MimicBERT finetuned using traditional NER task. BioBERT and MimicBERT are the current state-of-the-art (SOTA) models for NER on multiple biomedical datasets. The second set of baselines are BioBERT, MimicBERT and BERT-MRC finetuned on the knowledge guided NER task. BERT-MRC is initialized with BioBERT base weights, same as our BERT-CNN model. BERT-MRC model is another concurrent query-driven NER model \cite{li2019unified}, that models the task as a machine reading comprehension task. It predicts all possible start and end positions and predicts valid start-end spans through another feed-forward layer that takes input the predicted start-ends. This model shows considerable improvements in general domain query based NER tasks. The baselines are trained on individual datasets as each dataset has a separate set of entities. 

\section{Results and Discussion}
\subsection{Biomedical NER}
\label{results}
Table \ref{tab:sota} compares our method with our baselines on the 18 biomedical NER datasets. Our methods use the best knowledge context identified on the validation set performance.
Current state-of-the-art for AnatEM and Linnaeus use specific lexicons and entity specific rules that do not generalize, and hence are not directly comparable to neural models, although our methods approach their performance. On BC4CHEMD and BC5CDR the state-of-the-art methods are BERT-Base models finetuned specifically on chemical and other science corpus, whereas our methods use BioBERT as backbone. Still our models are within 1\% F1 score. 2011-i2b2 does not have a task specific to NER, therefore does not have current state-of-the-art methods, but still has annotations for the named entities which we use for joint training. On the rest 12 datasets, we achieve state-of-the-art using BERT-Base and beat methods that use BERT-Large. On JNLPBA and NCBI-Disease datasets we improve, but our improvement is not statistically significant. The SOTA scores are F1 values from the following work \cite{crichton2017neural,si2019enhancing,lee2019biobert,beltagy2019scibert}.

Our task formulation gives a significant boost in performance, which is observed in the improvements made by our BioBERT and MimicBERT base models compared to the baseline models following traditional NER formulation. Our BERT-CNN model further improves performance over BioBERT knowledge guided QA model on 12 tasks with a margin of 0.5 to 2.2\% (173 - 1934 samples). When it under-performs, it is within a margin of 0.5\% (less than 100 samples).  

BERT-MRC fails to perform strongly using the same knowledge context as BERT-CNN. On analysis, we observe the model fails to predict the correct end locations for majority of the samples. Overall, BERT-MRC suffers in Recall. When we compare our BIO tagging scheme to BERT-MRC start-end prediction method, if $k$ is the number of entities and $N$ is the number of tokens, time complexity wise our method is $\mathcal{O}(kN)$ as we classify each token, whereas BERT-MRC is $\mathcal{O}(kN^2)$ as they independently predict start and end locations and then match each start location with an end location.

\subsection{Ablation Studies and Analysis}

\begin{table}[t]
        \centering
        \scriptsize
        \resizebox{1.0\columnwidth} {!}{
        \begin{tabular}{lcccccc}
            \toprule 
            \multirow{1}{*}{\textbf{DATASET}}
            &  \multicolumn{1}{c}{\textbf{P}}
            &  \multicolumn{1}{c}{\textbf{$\Delta$P }}
            & \multicolumn{1}{c}{\textbf{R}}
            &  \multicolumn{1}{c}{\textbf{$\Delta$R}}
            & \multicolumn{1}{c}{\textbf{F}}
            &  \multicolumn{1}{c}{\textbf{$\Delta$F}}\\
             \hline
             \multirow{1}{*}{
                \rotatebox[origin=c]{0}{
                    \shortstack{ANATEM}}}
            &87.34 &\textbf{-1.47}&  89.31 &{+0.19}& 88.31 &\textbf{-0.65}\\

             \multirow{1}{*}{
                \rotatebox[origin=c]{0}{
                    \shortstack{BC2GM}}}
            &79.91 &\textbf{-2.9}8&  81.63 &\textbf{-1.76}& 80.76 &\textbf{-2.38} \\ 
             
            \multirow{1}{*}{
                \rotatebox[origin=c]{0}{
                    \shortstack{BC4CHEMD}}}
            &92.56 &{+0.49}&  91.10 &{+0.09}& 91.82 &{+0.28}\\ 
             
             \multirow{1}{*}{
                \rotatebox[origin=c]{0}{
                    \shortstack{BC5CDR}}}
            &87.67 &\textbf{-2.42}&  90.13 &{+0.97}& 88.88 &\textbf{-0.74}\\
             
              \multirow{1}{*}{
                \rotatebox[origin=c]{0}{
                    \shortstack{BIONLP09}}}
            &86.92 &\textbf{-3.24}&  84.91 &\textbf{-6.61}& 85.90 &\textbf{-4.93}\\
             
             \multirow{1}{*}{
                \rotatebox[origin=c]{0}{
                    \shortstack{BIONLP11EPI}}}
            &84.57 &\textbf{-4.01}&  86.97 &\textbf{-0.43}& 85.75 &\textbf{-2.24}\\

             \multirow{1}{*}{
                \rotatebox[origin=c]{0}{
                    \shortstack{BIONLP11ID}}}
            &87.98 &{+1.38}&  84.64 &\textbf{-0.71}& 86.27 &{+0.30}\\

              \multirow{1}{*}{
                \rotatebox[origin=c]{0}{
                    \shortstack{BIONLP13CG}}}
            &84.01 &\textbf{-3.97}&  80.84 &\textbf{-6.42}& 82.39 &\textbf{-5.23}\\
             
             \multirow{1}{*}{
                \rotatebox[origin=c]{0}{
                    \shortstack{BIONLP13GE}}}
            &72.33 &\textbf{-9.49}&  86.44 &{+0.18}& 78.76 &\textbf{-5.22}\\
             
             \multirow{1}{*}{
                \rotatebox[origin=c]{0}{
                    \shortstack{BIONLP13PC}}}
            &86.40 &\textbf{-2.63}&  87.21 &\textbf{-4.66}& 86.80 &\textbf{-3.63}\\
             
             \multirow{1}{*}{
                \rotatebox[origin=c]{0}{
                    \shortstack{CRAFT}}}
            &86.35 &\textbf{-1.72}&  85.65 &\textbf{-2.54}& 86.00 &\textbf{-2.13}\\
             
             \multirow{1}{*}{
                \rotatebox[origin=c]{0}{
                    \shortstack{EXPTM}}}
            &75.28 &\textbf{-8.44}&  81.51 &\textbf{-4.23}& 78.27 &\textbf{-6.44}\\
             
              \multirow{1}{*}{
                \rotatebox[origin=c]{0}{
                    \shortstack{JNLPBA}}}
            &76.85 &{+0.81}&  81.79 &{+0.16}& 79.24 &{+0.51}\\
             
             \multirow{1}{*}{
                \rotatebox[origin=c]{0}{
                    \shortstack{LINNAEUS}}}
            &91.28 &{+2.81}&  86.15 &\textbf{-2.32}& 88.64 &{+0.17}\\
             
             \multirow{1}{*}{
                \rotatebox[origin=c]{0}{
                    \shortstack{NCBI-DISEASE}}}
            &83.86 &\textbf{-2.80}&  87.25 &\textbf{-3.63}& 85.52 &\textbf{-3.20}\\
             
             \multirow{1}{*}{
                \rotatebox[origin=c]{0}{
                    \shortstack{2010-i2b2}}}
            &89.87 &\textbf{-5.40}&  90.75 &\textbf{-5.16}& 90.31 &\textbf{-5.28}\\

             \multirow{1}{*}{
                \rotatebox[origin=c]{0}{
                    \shortstack{2011-i2b2}}}
            &91.49 &\textbf{-2.93}&  92.25 &\textbf{-2.12}& 91.87 &\textbf{-2.53}\\ 

            \multirow{1}{*}{
                \rotatebox[origin=c]{0}{
                    \shortstack{2012-i2b2}}}
            &82.05 &{+0.72}&  82.31 &\textbf{-2.21}& 82.18 &\textbf{-0.71}\\
             \hline

            \end{tabular}
            }
            \caption{
            Change in performance when BERT-CNN model is trained individually on respective datasets with \textbf{Question} context. Negative $\Delta$ indicates Multi-task is better and are in bold.
            Precision (P), Recall (R), F-Measure (F). 
            }
            \label{tab:abl_mt}
        \end{table}

\begin{figure}[t]
  \includegraphics[width=\linewidth]{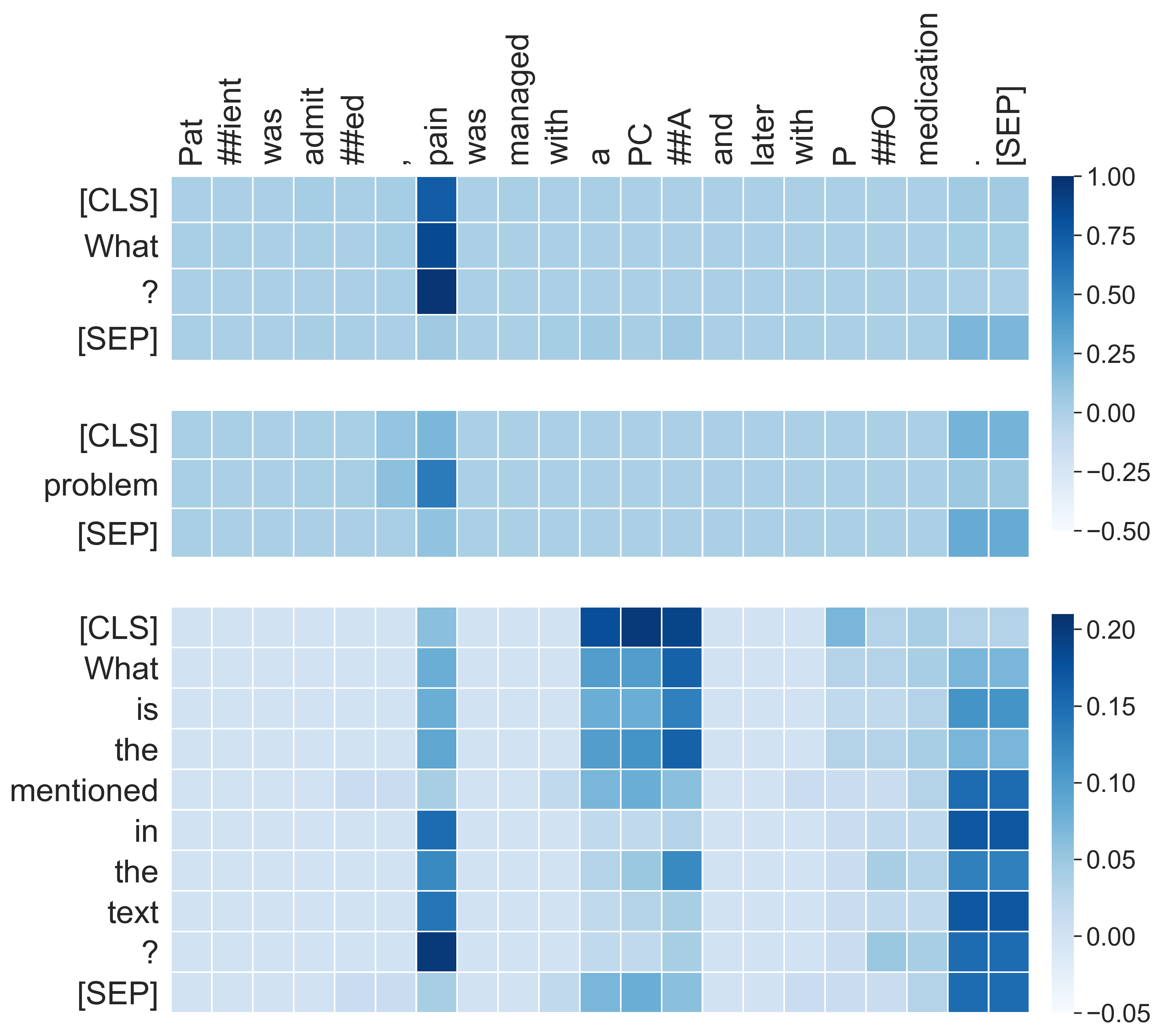}
  \caption{Attention scores of our BERT-CNN model trained with \textit{Question} context for three knowledge context probes, ``What'', ``problem'' and a context where we remove the entity-type. For the last, the model attends to all three entities present \textbf{pain} (\textit{problem}), \textbf{PCA} and \textbf{PO medication} (\textit{treatment}) }
  \label{fig:viz}
\end{figure}

\begin{figure*}[t]
  \includegraphics[width=\textwidth,height=3.8cm]{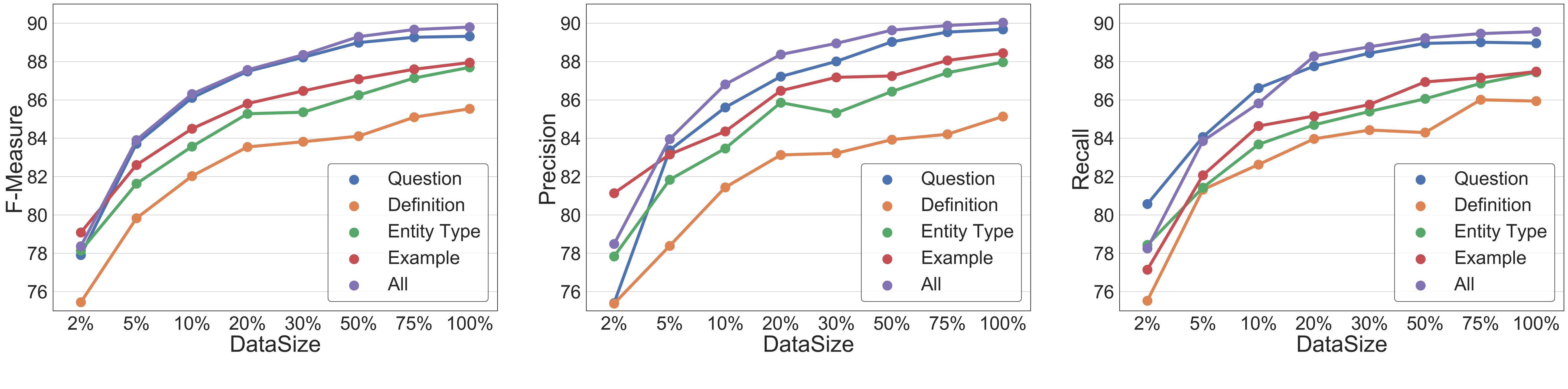}
  \caption{ 
  Effect of the train set size on the three validation set metrics for BERT-CNN model with five contexts.
  }
  \label{fig:mtlearning}
\end{figure*}

\paragraph{Effect of different knowledge contexts:}

Since we incorporate four different knowledge contexts to help in NER, here we identify which knowledge context is better for the NER task across all the 18 datasets.  
The performance of BERT-CNN model with the knowledge contexts across the test set is shown in Table \ref{tab:knowledge}. The scores are entity precision, recall and exact match F1 scores. We observe \textit{Question} and \textit{All} contexts to perform consistently on all the datasets. We believe this is because of the presence of ``what'' that helps the model to find entities much better than given just a text. To verify our hypothesis we probe our BERT-CNN model trained with \textit{Question} context with multiple probes like ``what problem?'', the complete question, and ``problem'' for 20 samples and observe the change in attention scores. As the model is trained with a template, the best prediction is observed on the complete question, with the least scores for only the entity type. Attention scores for ``what'' were consistently high. The question of why BERT attention scores are high for query words is a future research problem, and similar questions about Attention have been raised by others \cite{clark2019does}. 
Figure \ref{fig:viz} shows examples of such probes. We can observe our model identifies all the entities, and the entity-type acts as a filter. \textit{Definition} and \textit{Examples} under-perform, we believe the definition might be too generic for an entity type and examples, although representative of class might not be comprehensive. 

\begin{table}[t]
\centering
\scriptsize
\resizebox{\columnwidth} {!}{
\begin{tabular}{lcccccc}
\toprule
\multirow{2}{*}{\begin{tabular}[c]{@{}l@{}}\textbf{ENTITY}$\downarrow$ \end{tabular}} & \multicolumn{2}{c}{\textbf{PRECISION}} & \multicolumn{2}{c}{\shortstack{\textbf{RECALL}}} & \multicolumn{2}{c}{\textbf{F-MEASURE}} \\ \cline{2-7} 

& \textbf{No-K}            & \textbf{K}            &\textbf{No-K}           & \textbf{K}          & \textbf{No-K}            & \textbf{K}           \\ \hline
Gene/Protein    & 67.33 & \textbf{84.19} & 74.01 & \textbf{86.73} & 70.51 & \textbf{85.44}\\ 
Chemical      &  89.04 & \textbf{91.84} & 88.42 & \textbf{91.15} & 88.73 & \textbf{91.49}\\ 
Disease         & 83.17 & \textbf{83.48} & 86.62 & \textbf{88.00} & 84.86 & \textbf{85.68}\\
Problem     &91.95 & \textbf{93.28} & 92.83 & \textbf{94.18} & 92.39 & \textbf{93.73}\\ 
Treatment      &  91.78 & \textbf{92.91} & 91.98 & \textbf{93.47} & 91.88 & \textbf{93.19}\\
Test          &    92.12 & \textbf{94.09} & 93.23 & \textbf{94.67} & 92.67 & \textbf{94.38}\\\hline
\end{tabular}
}
\caption{
Comparison of Entity specific (No-K) BERT-CNN model with Question Context provided multi-entity BERT-CNN model (K). Better values are in bold}
\label{tab:kvsnok}
\end{table}

\noindent
\textbf{Effect of Multi-task training:}
To analyze how much multi-task training strategy affects the performance, we define an experiment where we keep the model (BERT-CNN) and the knowledge context \textit{Question} same, but train on each individual datasets, and compare with joint training. Table \ref{tab:abl_mt} shows the results of our experiments. Individual dataset training helps improving F1 scores marginally (max 0.5\%) on four datasets, whereas joint training substantially improves performance (0.65\%-6.44\%) on some datasets. This empirically validates our hypothesis that training on combined huge biomedical datasets helps. 
On further analysis the datasets that have unique entities like AnatEM, 2012-i2b2, Bionlp11ID, Linnaeus and JNLPBA do not show much difference. The datasets that improve is due to presence of common entity-types like Gene/Protein, Chemical, Disease, Problem, Treatment and Test that helps the model to generalize well and learn better representations. We noticed that the definition of an entity was consistent when it was present in multiple datasets. It ensured that the model was not confused by different definitions. On the other hand, some identical entities had different names in different datasets. For example, disease was called Problem in i2B2, Disease in NCBI-Disease, and by a more specific name Cancer in Bionlp13CG. 

\noindent
\textbf{Effect of Knowledge Context compared to Individual entity training:}
In this experiment we study the effect of knowledge context over our task formulation. We compare our single BERT-CNN model trained with \textit{Question} context to six different BERT-CNN models each trained only for one specific entity type detection without any context. The entity types are selected such that they are present in multiple datasets and have the most significant number of samples. Table \ref{tab:kvsnok} summarizes the results. The results show that knowledge context and training jointly with multiple entity types helps in improving the performance for all entity types, compared to individual entity specific models. The performance improvement for Gene/Protein is the most significant. 

\noindent
\textbf{Effect of Train Set Size:}
In this experiment we study how the train set size affects the model performance.  We sample different percentage of training samples from the total train dataset as seen in Figure \ref{fig:mtlearning}. We ensure a balanced sampled train set with equal number of positive and negative samples and evaluate across all entities. The training samples are chosen from each of the datasets to ensure the model is not biased towards a dataset or entity type. We do not change any parameters of the model.  
It can be observed that the performance of the model increases rapidly and then tapers down for each of the context types. We can infer that the model can achieve quite a good performance (84\%) with just 5\% of training samples but needs much more samples to achieve the state-of-the-art performance. As training samples goes beyond 5\%, the precision, recall and the F1-scores for knowledge types \textit{Question} and \textit{All} clearly separate themselves from the other contexts. 

\begin{table}[t]
\centering
\scriptsize
\resizebox{1.0\columnwidth} {!}{
\begin{tabular}{lcccccc}
\toprule
\multirow{2}{*}{\textbf{\begin{tabular}[c]{@{}l@{}}MODEL $\xrightarrow{}$\\ ENTITY $\downarrow$\end{tabular}}} & \multicolumn{2}{c}{\textbf{BIOBERT(Ours)}} & \multicolumn{2}{c}{\textbf{MimicBERT(Ours)}} & \multicolumn{2}{c}{\textbf{BERT-CNN(Ours)}} \\ \cline{2-7} 
                                                                                      & \textbf{SRC F1}   & \textbf{TGT F1}   & \textbf{SRC F1}    & \textbf{TGT F1}    & \textbf{SRC F1}   & \textbf{TGT F1}   \\ \hline
Gene/Protein                                                                          & 84.83             & \underline{83.27}             & 82.46              & 80.75              & 84.99             & \textbf{85.63}             \\ 

Chemical                                                                              & 91.35             & \textbf{76.13}             & 85.60              & 66.63              & 89.83             & \underline{75.92}             \\ 
Disease                                                                               & 86.46             & \underline{68.01}             & 84.13              & 60.54              & 88.74             & \textbf{70.00}             \\ 
Problem                                                                               &  94.42            & \underline{90.29}             &  93.72             & 89.67              & 94.43             & \textbf{90.90}             \\ 
Treatment                                                                             &94.01             & 89.67             & 93.76              & \underline{89.99}              & 94.16             & \textbf{90.22}             \\ 
Test                                                                                  & 94.99             & \textbf{91.36}             &  94.85             & 90.91              &  94.85             & \underline{91.11}            \\ \hline

\end{tabular}
}
\caption{
Transfer Learning experiment results. The metric is exact match F1 for source (SRC) and target (TGT) domain. Bold across each of the entities are the best, underlined are the second best.
    }
\label{tab:tl}
\end{table}

\noindent
\textbf{Transfer Learning:}
We also examine the transfer learning capability of our system on the six entity sets: gene/protein, chemical, disease, problem, test and treatment, since they are present in multiple datasets. We tested our model on samples of one dataset (target domain) while training and validating on the remaining datasets (source domain).  
We choose the target domain for each entity to be the dataset that produces the best overall F1-score on full data with BERT-CNN model.  
We consider Bionlp11ID, BC4CHEMD, BC5CDR datasets as the target domain for gene/protein, chemical and disease respectively and 2010-i2b2 for problem, treatment and test. Table \ref{tab:tl} summarizes the results. 

The results show a varied degree of transfer learning, losing only 0.5\% F1 in some tasks and by as much as 20\% in other tasks, which is in line with earlier observed performance loss \cite{bethard-etal-2017-semeval}. 
The difference in source and target F1-scores is remarkably close for Problem, Treatment and Test entities. The two domains although are close for these entities, they do have different set of entities. Chemical shows a significant drop but still our model achieves 75\% F1 despite the target domain containing many prior unseen entities. For Gene/Protein the source and target domain are nearly the same set of entities. 

\section{Related Work}
\textbf{External Knowledge} : In the past, there have been several attempts to incorporate external knowledge through feature engineering and lexicons \cite{liu-etal-2019-towards,borthwick1998exploiting,ciaramita2005named,kazama2007exploiting}, or incorporating knowledge in the feature extraction stage \cite{crichton2017neural,yadav-bethard-2018-survey}, or using document context \cite{devlin2018bert}.
In our work, we incorporate simple textual knowledge sentences and show how to integrate them in named entity recognition tasks.

\noindent
\textbf{Multi-Task Learning} : Multi-task learning have been used in the past to tackle the labelling problem of NER. For example, multi-task learning with simple word embedding and CNN \cite{crichton2017neural}, cross-type NER with Bi-LSTM and CRF \cite{wang2018cross},  MTL with private and shared Bi-LSTM-CRF using character and word2Vec word embeddings \cite{wang2019multitask}. In our work, we do multi-task learning by reducing all different NER tasks to the same generic format.

\noindent
\textbf{Language Models and Transfer Learning} : There have been prior attempts to reduce the labelling confusion by using a single model to predict each entity-type \cite{lee2019biobert} and also using transfer-learning \cite{lee2019biobert, beltagy2019scibert, si2019enhancing}. Our work is similar to them, which also use pre-trained language models (BERT), and/or predict different types of entities separately, but differs in task formulation and use of explicit external knowledge context. We show jointly learned single model is better than per entity-type models.

\noindent
\textbf{NER as a Question Answering Task} : In general domain, researchers have formulated multiple NLP tasks as question-answering format in DecaNLP \cite{mccann2018natural}, semantic-role labelling as in QASRL \cite{he2015question} and others have argued that question-answering is a format not a task \cite{gardner2019question}.  We also use QA format as a part of our task, to address previously mentioned challenges . A possibly concurrent work, BERT-MRC \cite{li2019unified} also attempts at NER as a QA task in general domain by span predictions for individual entity types in a reading comprehension style approach. We however differ in the task formulation using BIO tagging scheme, our model design and our focus in Biomedical NER. A detailed comparison is in Section \ref{results}.

\section{Conclusion}
We reformulated the NER task as a knowledge guided, context driven QA task and showed it has a significant impact. Our models are more explainable using the query-text attention and address the major challenges faced by current NER systems. Our approach has achieved above state-of-the-art F measures for 14 of the common public biomedical NER datasets.  In future, we plan to perform more experiments, such as few-shot learning between different entity groups, adding specific loss functions and logical constraints for NER tasks. 

\section*{Acknowledgments}

The authors acknowledge the support from the DARPA SAIL-ON program, and ONR award N00014-20-1-2332.

\bibliography{anthology,emnlp2020}
\bibliographystyle{acl_natbib}

\newpage
\appendix
\section{Supplemental Material}
\label{sec:supplemental}


\subsection{Model Explanation with Attention Probing}
We study our BERT-CNN model using  attention value heatmaps and try to explain how our model uses the knowledge contexts to extract specific entities from a given dataset. To show this,  we choose a sample ``\textit{Patient} was \textit{admited} , \textit{pain} was managed with \textit{a PCA} and later with \textit{PO medication}.'' where there are multiple entities \textbf{pain} (problem), \textbf{a PCA} and \textbf{PO medication} (treatment), \textbf{Patient} (person) and \textbf{admitted} (occurance).

\begin{figure}[ht]
  \includegraphics[width=\linewidth]{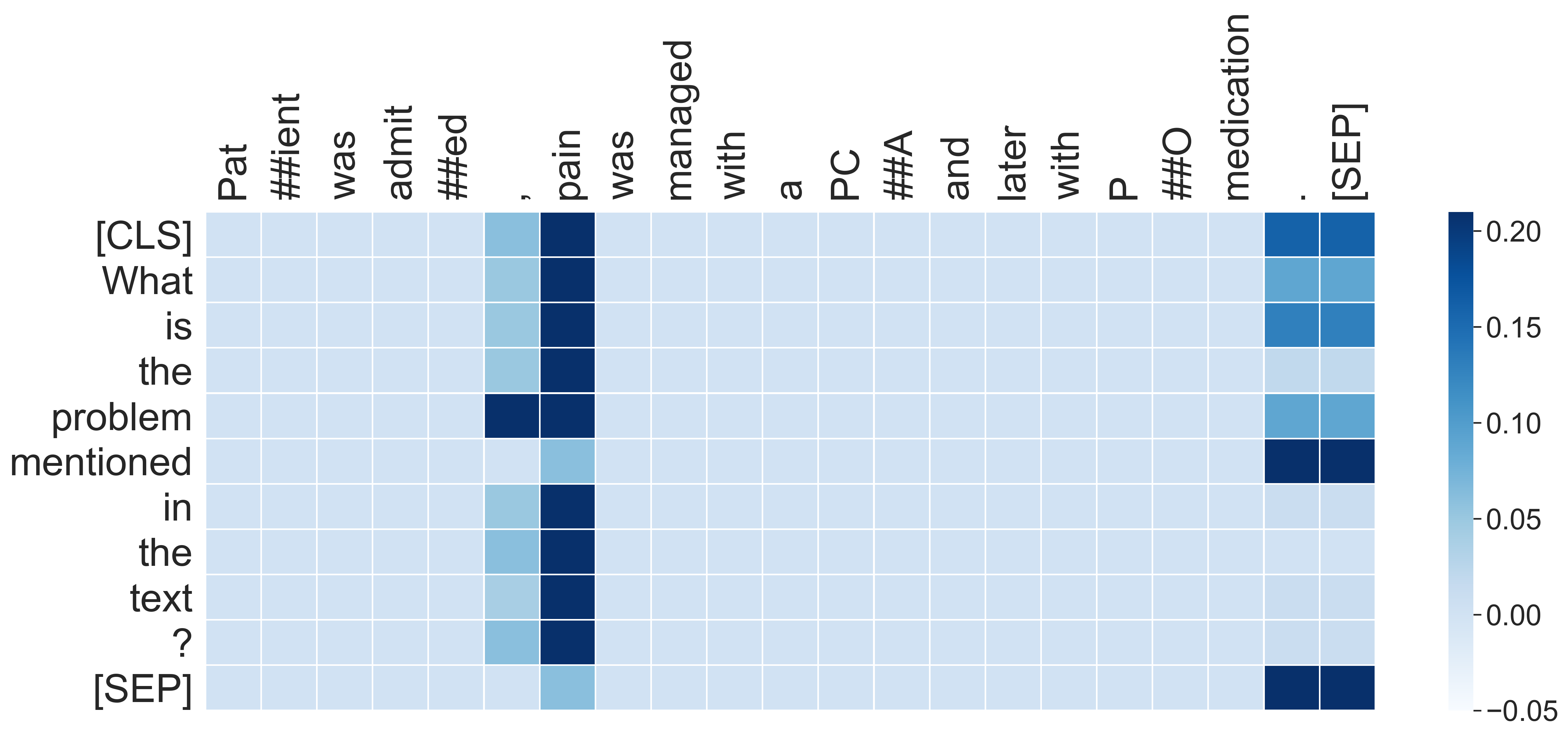}
  \caption{Problem - Question}
  \label{fig:problemQ}
\end{figure}

\begin{figure}[ht]
  \includegraphics[width=\linewidth]{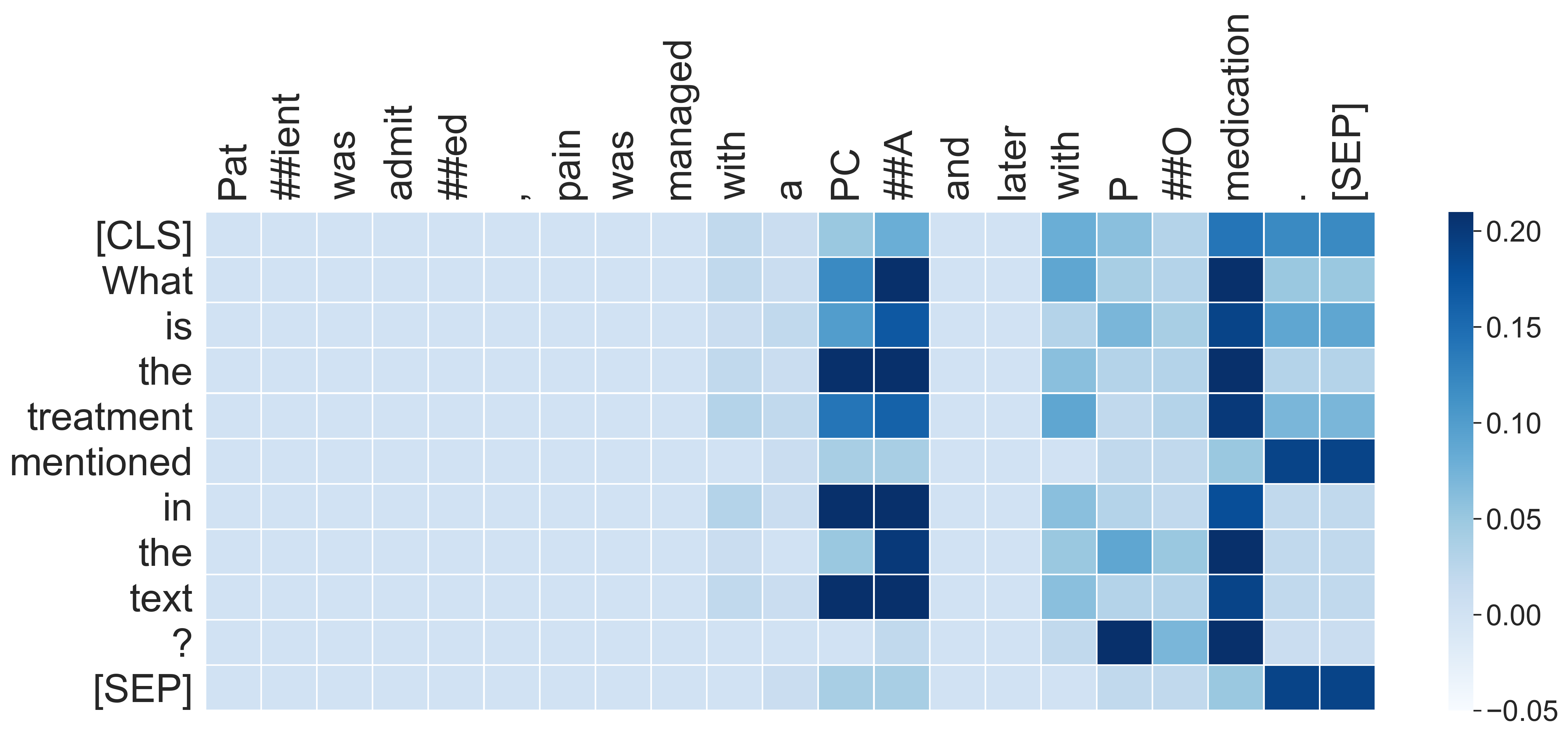}
  \caption{Treatment - Question}
  \label{fig:treatmentQ}
\end{figure}
\subsubsection{Using Question as Knowledge Context}
\begin{figure}[ht]
  \includegraphics[width=\linewidth]{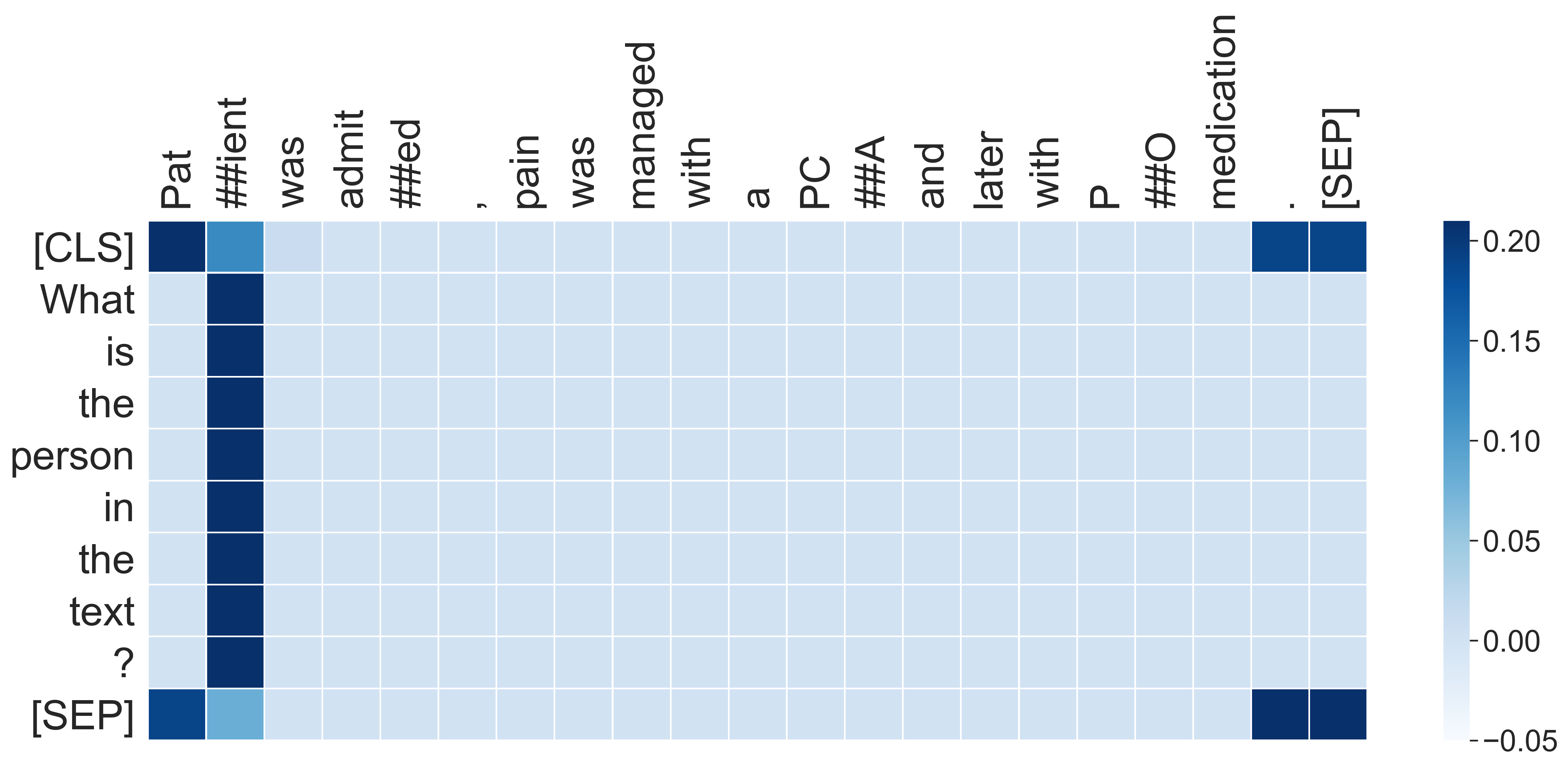}
  \caption{Person - Question}
  \label{fig:personQ}
\end{figure}

\begin{figure}[ht]
  \includegraphics[width=\linewidth]{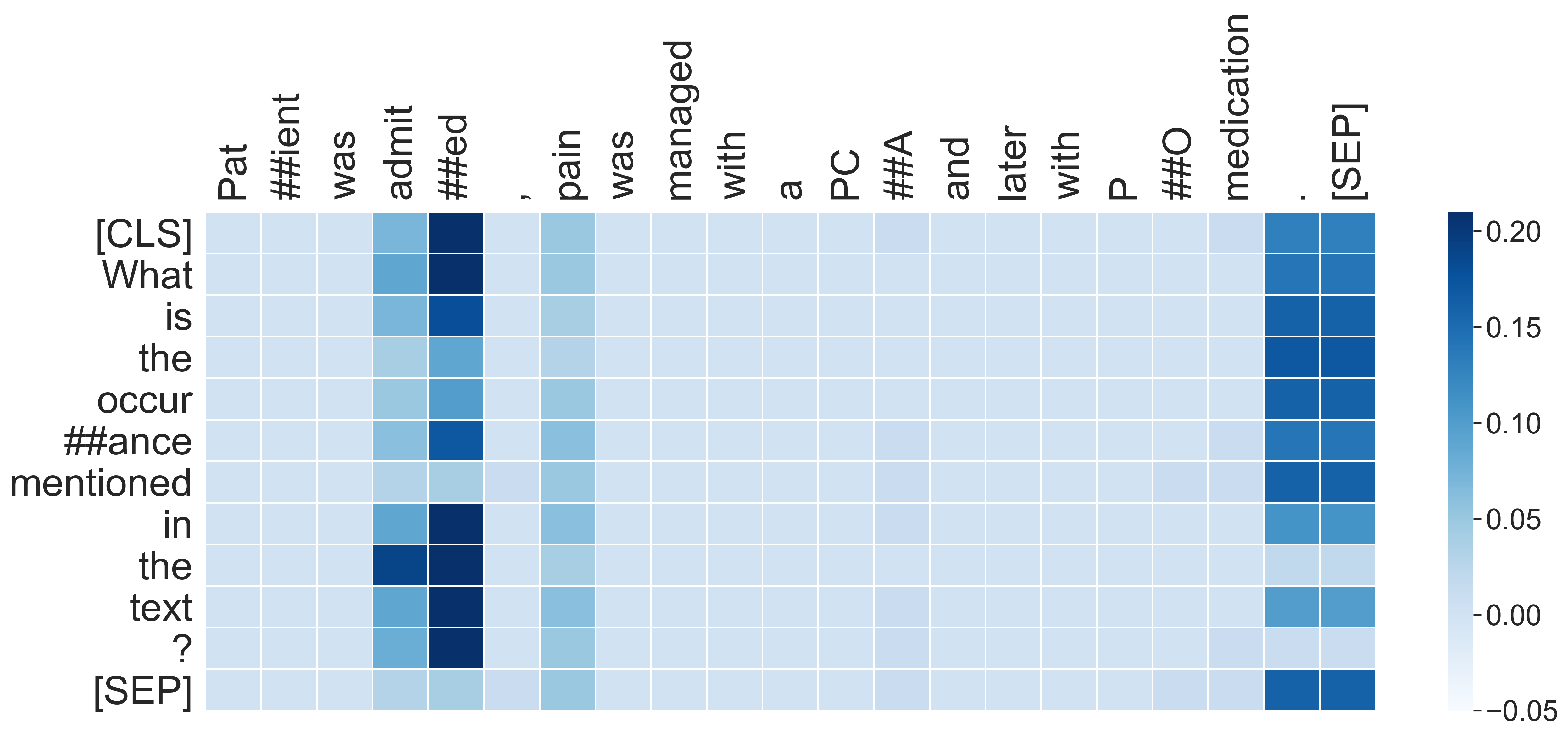}
  \caption{Occurance - Question}
  \label{fig:occuranceQ}
\end{figure}

From the Figures \ref{fig:problemQ}, \ref{fig:treatmentQ}, \ref{fig:personQ} and \ref{fig:occuranceQ} it can be seen that, when knowledge context is in question format then, each of the entity types present in the knowledge context guide the model to chose the correct entities. This can be seen from the higher attention values for those specific entities.

\subsubsection{Using Entity type, Definition, Example and All combined as Knowledge Context}
\begin{figure}[ht]
  \includegraphics[width=\linewidth]{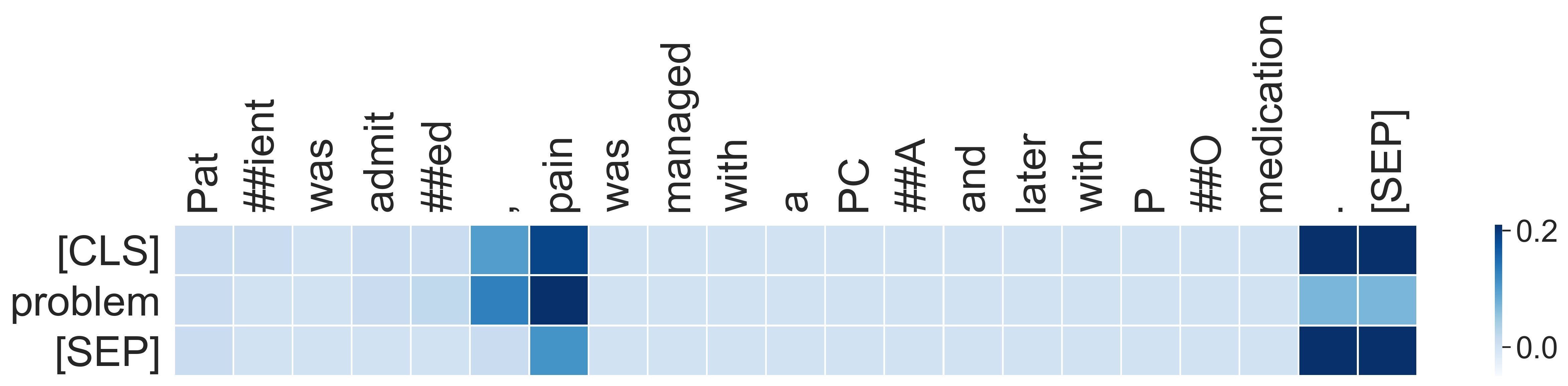}
  \caption{Problem - Entity Type}
  \label{fig:problemT}
\end{figure}

\begin{figure}[ht]
  \includegraphics[width=\linewidth]{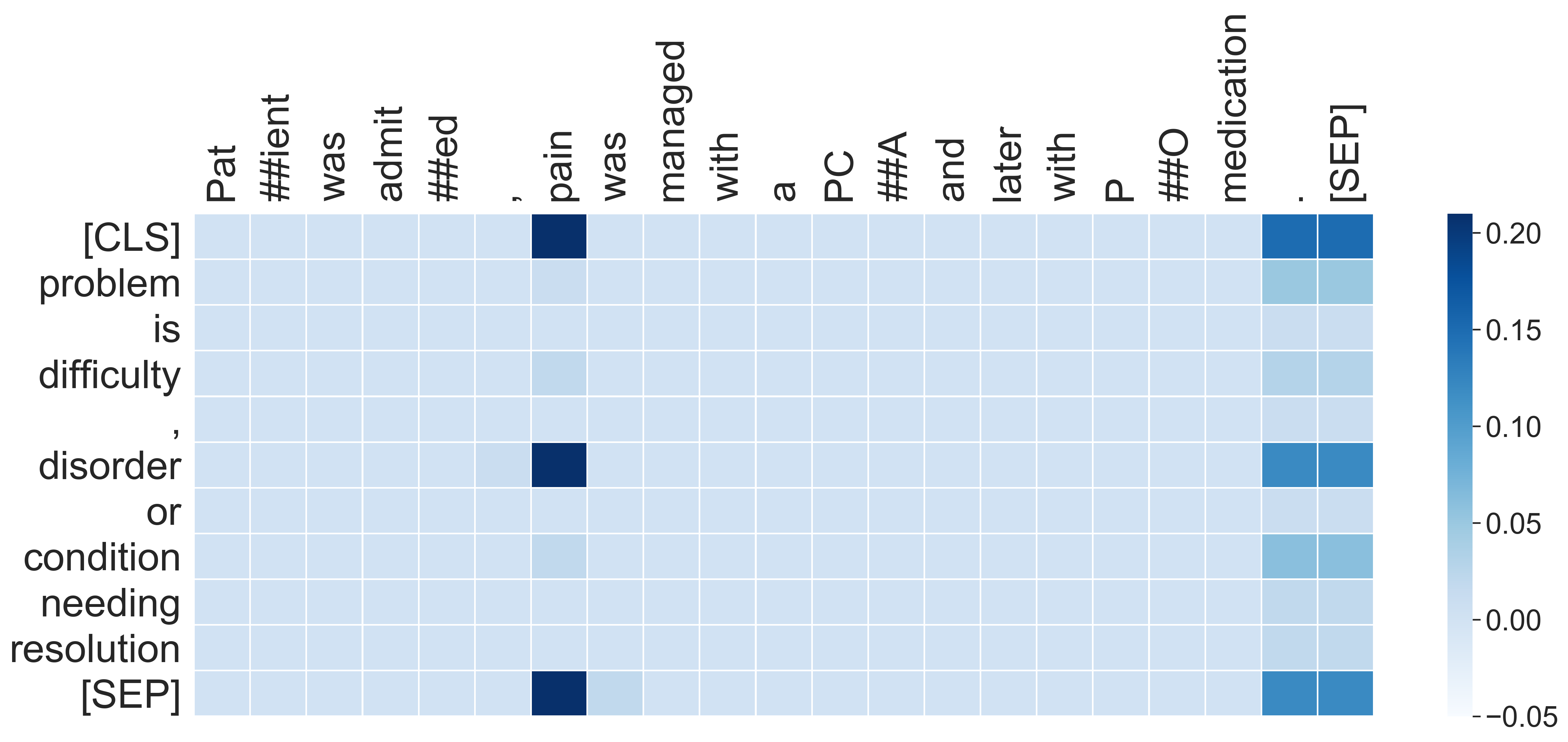}
  \caption{Problem - Definition}
  \label{fig:problemD}
\end{figure}

\begin{figure}[ht]
  \includegraphics[width=\linewidth]{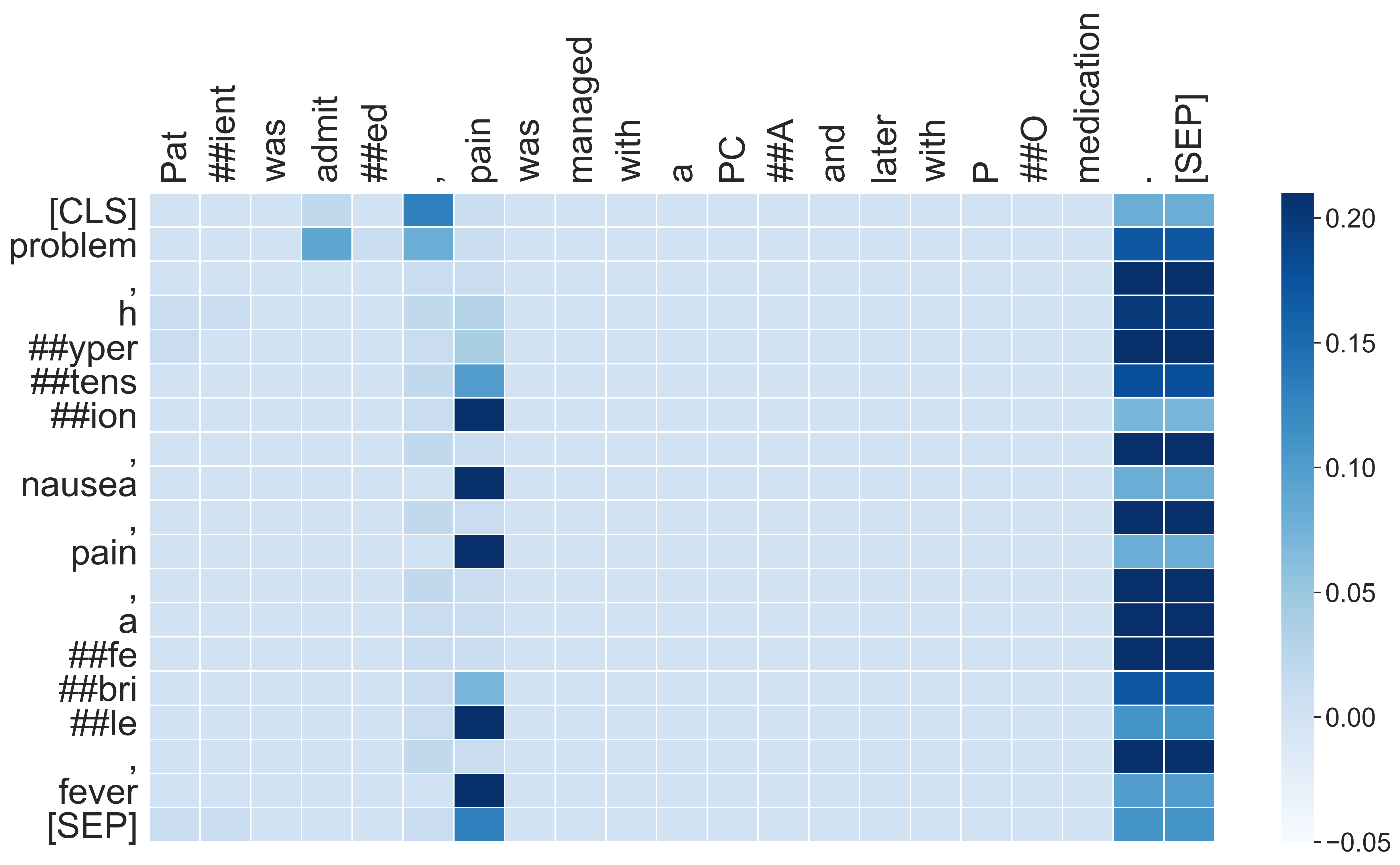}
  \caption{Problem - Example}
  \label{fig:problemE}
\end{figure}

\begin{figure}[ht]
  \includegraphics[width=\linewidth]{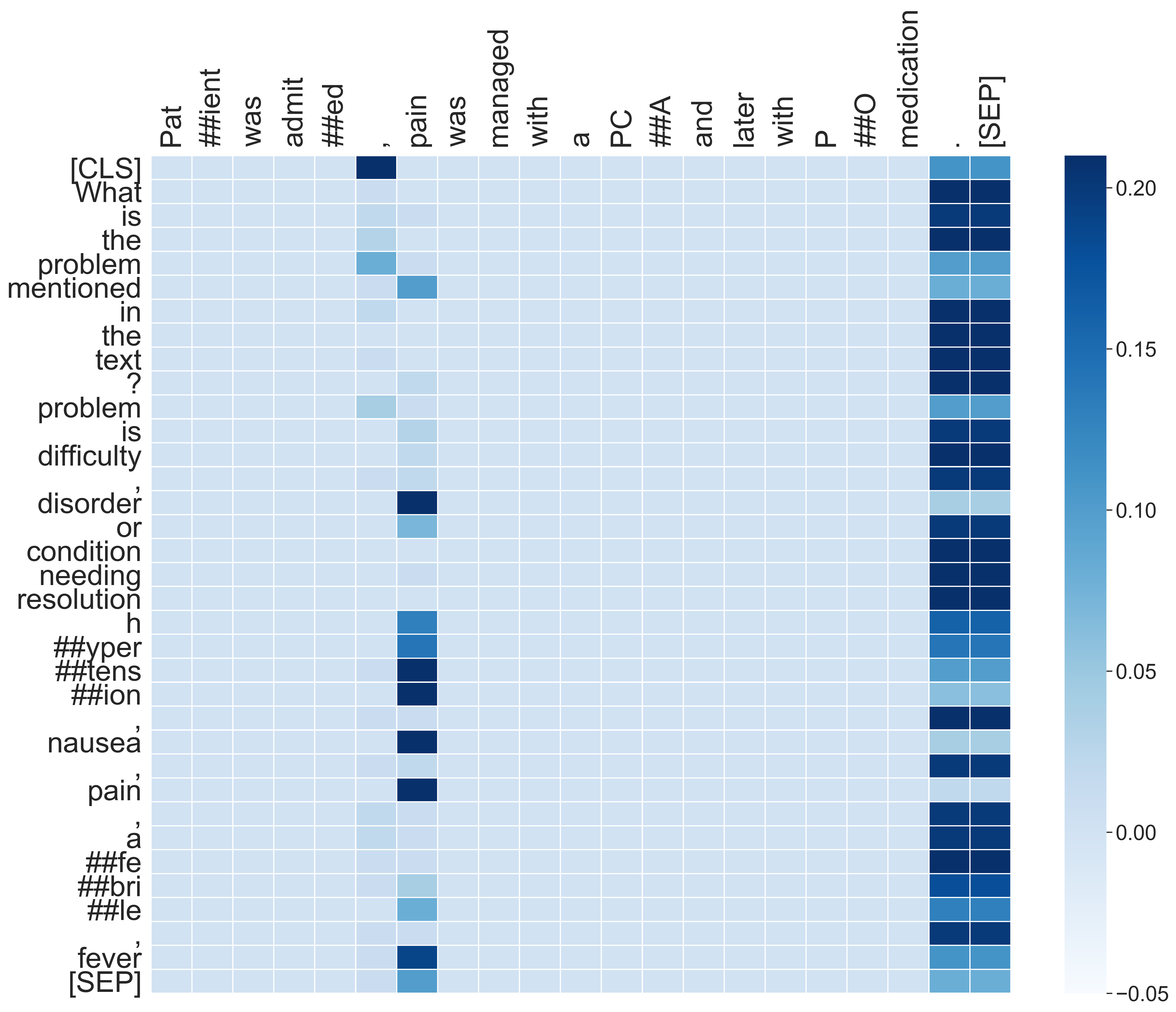}
  \caption{Problem - All knowledge combined}
  \label{fig:problemA}
\end{figure}

We also probed our model to extract the attention weights for each of the other four knowledge contexts. Here we show this only for problem entity type. In Figure \ref{fig:problemT} it can be seen that attention weight between \textit{problem} in knowledge context and \textit{pain}  in text is highest. Figure \ref{fig:problemD} shows keyword like \textit{disorder} in the definition representing the meaning of the entity type problem highly attends to the entity \textit{pain}. On using  the example as  knowledge, it can be seen from Figure \ref{fig:problemE}, keywords like hypertension, pain, nausea, fever highly attend to the entity pain in the text. This is in line with our hypothesis that providing similar entities belonging to the same entity group might help find the entity in a text. Finally, in Figure \ref{fig:problemA}, it can be seen that the keywords from question, definition and example all collectively help to predict the entity pain in the text.

\subsection{Model Training}
We use the HuggingFace \cite{Wolf2019HuggingFacesTS} and Pytorch Deep learning framework \cite{NEURIPS2019_9015}.
We train the model with following hyperparameters, learning rates in the range [1e-6,5e-5], batch sizes of [16,32,48,64], linear weight-decay in range [0.001,0.1] and warm-up steps in range of [100,1000]. We use BERT-base-cased version for all our models. The BERT-base-cased model has  nearly  110M parameters.

\subsection{Entity Distribution in the Dataset}
The number of samples present in the Table \ref{tab:data} are created directly from the train, validation and test samples of 18 biomedical datasets. The  i2b2 datasets do  not have separate validation data splits.  We use 30\% of the samples from training data as validation data.
The \textit{Entity Mentions} represents the total number of entities  present for the datasets in all of train,  validation and test samples. Since each sample data can have multiple entities, the number is higher than the total positive and negative samples for the dataset.

 \subsection{Performance comparison on  Test Data}
 Table \ref{tab:sota_2} shows the performance comparison of our best model with SOTA on the test data for each of the 18 datasets.
  The F-measures in bold are the best performance on each datasets. The state-of-the-art for Linnaeus and AnatEM datasets uses dictionaries developed without a clear train/test split, hence our scores are not directly comparable. Also 2011-i2b2 do not have SOTA concept extraction performance. Improvements in 10 datasets  are significant as   compared  to the SOTA.

\begin{table}[ht]
        \centering
        \small
        \scriptsize
        \resizebox{\columnwidth} {!}
        {
        \begin{tabular}{l r c c c c c}
            \toprule 
            \textbf{Dataset} &  
            \multicolumn{1}{c}{\textbf{Entities}} & \multicolumn{1}{c}{\textbf{SOTA F1 }} & \multicolumn{1}{c}{\textbf{OURS P }} & \multicolumn{1}{c}{\textbf{OURS R}} & \multicolumn{1}{c}{\textbf{OURS F1}}  & \multicolumn{1}{c}{\textbf{Significant}}\\
            \hline
            ANATEM & 4616 & \textbf{91.61} & 90.29 & 89.43 & 89.85 $\pm$ 0.48 & No \\ 
BC2GM & 6322 & 81.69 & 82.89 & 83.39 & \textbf{83.14 $\pm$ 0.54} & Yes \\ 
BC4CHEMD & 25331 & \textbf{92.36} & 92.56 & 91.10 & 91.82 $\pm$ 0.58 & No \\ 
BC5CDR & 9808 & \textbf{90.01} & 90.09 & 89.62 & 89.62 $\pm$ 0.71 & No \\ 
BIONLP09 & 3589 & 84.20 & 91.55 & 92.95 & \textbf{92.25 $\pm$ 0.57} & Yes \\ 
BIONLP11EPI & 5730 & 78.86 & 88.58 & 87.40 & \textbf{87.99 $\pm$ 1.10} & Yes \\ 
BIONLP11ID & 3810 & 81.73 & 87.98 & 84.64 & \textbf{86.27 $\pm$ 1.80} & Yes \\ 
BIONLP13CG & 7861 & 78.90 & 90.62 & 88.56 & \textbf{89.58 $\pm$ 0.68} & Yes \\ 
BIONLP13GE & 4354 & 78.58 & 83.77 & 88.01 & \textbf{85.84 $\pm$ 0.93} & Yes \\ 
BIONLP13PC & 5306 & 81.92 & 90.14 & 92.09 & \textbf{91.11 $\pm$ 0.12} & Yes \\ 
CRAFT & 18770 & 79.56 & 90.54 & 89.19 & \textbf{89.86 $\pm$ 0.55 }& Yes \\ 
EXPTM & 2308 & 74.90 & 85.97 & 85.30 & \textbf{85.64 $\pm$ 0.61} & Yes \\ 
JNLPBA & 8673 & 78.58 & 76.85 & 81.79 & \textbf{79.24  $\pm$  0.45} & No \\ 
LINNAEUS & 1428 & \textbf{95.68} & 90.69 & 90.53 & 90.61 $\pm$ 0.28 & No \\ 
NCBIDISEASE & 956 & 89.36 & 87.89 & 91.56 & \textbf{89.69 $\pm$ 0.37} & No \\ 
2010-i2b2 & 30140 & 90.25 & 95.27 & 95.91 & \textbf{95.59 $\pm$ 0.30} & Yes \\ 
2011-i2b2 & 25271 & - & 94.70 & 94.94 & 94.82 $\pm$ 0.41 & - \\ 
2012-i2b2 & 15301 & 80.91 & 84.83 & 85.25 & \textbf{85.04 $\pm$ 1.18} & Yes \\  \hline
            \end{tabular}
            }
            \caption{
Precision(P), Recall(R) and F-measure(F1) with our best model measured by running with three seed values. Significant column shows whether our F1-scores are  statistically significantly better than SOTA F1 ($p\leq0.05$, based on Wilson score intervals \cite{wilson1927probable}). Best F-measures are in bold.
                 }
            \label{tab:sota_2}
        \end{table}

\subsection{Training with balanced dataset}
We generated a sample  for each text available in the source data. The text may or may not contain a particular entity. So we generate negative samples for each text  and for each available entity types of the  dataset making the datasets unbalanced. In Table \ref{tab:sup_balanced} we show that the negative samples does not have much  impact  on the  performance of our models. The results are taken using BERT-CNN model with \textit{Question} as a context. Negative values in $\Delta P$,$\Delta R$ and $\Delta F$ means training on  unbalanced data is  better than on balanced data.


\subsection{Data Preprocessing}
We have done  the experiments on 18 biomedical dataset which are available in different formats. For the 15 publicly available datasets, we used the BIO annotated files and automatically extracted the spans based on the tags. The three  i2b2 files has different format. They have individual biomedical reports containing multiple sentences which may or may not contain the entities. So we preprocessed them by considering each statement as a sample without rejecting any  sentence. Thus we bring all the datasets  into a common format. Each sample in our pre-processed data contains the id(indicating the dataset which is the sample origin), text, answers, spans, number of answers present in the text, entity type, question context, definition of entities, top ten frequently occurring examples with counts. We also grouped together similar entity-types to form entity groups and add entity group definitions which can be used for further  research.

\begin{table}
        \centering
        \scriptsize
        \resizebox{1.0\columnwidth} {!}{
        \begin{tabular}{lcccccc}
            \toprule 
            \multirow{1}{*}{\textbf{DATASET}}
            &  \multicolumn{1}{c}{\textbf{P}}
            &  \multicolumn{1}{c}{\textbf{$\Delta$P}}
            & \multicolumn{1}{c}{\textbf{R}}
            &  \multicolumn{1}{c}{\textbf{$\Delta$R}}
            & \multicolumn{1}{c}{\textbf{F}}
            &  \multicolumn{1}{c}{\textbf{$\Delta$F}}\\
             \hline
             \multirow{1}{*}{
                \rotatebox[origin=c]{0}{
                    \shortstack{ANATEM}}}
            &88.88 &\textbf{-0.07}&  88.23 &{+0.89}& 88.55 &{+0.41}\\

             \multirow{1}{*}{
                \rotatebox[origin=c]{0}{
                    \shortstack{BC2GM}}}
            &82.93 &\textbf{-0.04}&  82.84 &{+0.55}& 82.88 &{+0.26}\\
             
            \multirow{1}{*}{
                \rotatebox[origin=c]{0}{
                    \shortstack{BC4CHEMD}}}
            &91.64 &{+0.43}&  91.03 &\textbf{-0.02}& 91.33 &{+0.21}\\ 
             
             \multirow{1}{*}{
                \rotatebox[origin=c]{0}{
                    \shortstack{BC5CDR}}}
            &89.61 &{+0.48}&  88.37 &{+0.79}& 88.98 &{+0.64}\\
             
              \multirow{1}{*}{
                \rotatebox[origin=c]{0}{
                    \shortstack{BIONLP09}}}
            &90.76 &\textbf{-0.60}&  92.33 &\textbf{-0.81}& 91.54 &\textbf{-0.71}\\
             
             \multirow{1}{*}{
                \rotatebox[origin=c]{0}{
                    \shortstack{BIONLP11EPI}}}
            &87.86 &{+0.72}&  86.29 &{+1.11}& 87.07 &{+0.92}\\

             \multirow{1}{*}{
                \rotatebox[origin=c]{0}{
                    \shortstack{BIONLP11ID}}}
            &85.39 &{+1.21}&  85.19 &{+0.16}& 85.29 &{+0.68}\\

              \multirow{1}{*}{
                \rotatebox[origin=c]{0}{
                    \shortstack{BIONLP13CG}}}
            &88.61 &\textbf{-0.63}&  87.54 &\textbf{-0.27}& 88.07 &\textbf{-0.45}\\
             
             \multirow{1}{*}{
                \rotatebox[origin=c]{0}{
                    \shortstack{BIONLP13GE}}}
            &81.94 &\textbf{-0.12}&  88.77 &\textbf{-2.51}& 85.22 &\textbf{-1.24}\\
             
             \multirow{1}{*}{
                \rotatebox[origin=c]{0}{
                    \shortstack{BIONLP13PC}}}
            &89.48 &\textbf{-0.45}&  90.73 &{+1.14}& 90.10 &{+0.33}\\
             
             \multirow{1}{*}{
                \rotatebox[origin=c]{0}{
                    \shortstack{CRAFT}}}
            &87.43 &{+0.64}&  88.12 &{+0.07}& 87.78 &{+0.35}\\
             
             \multirow{1}{*}{
                \rotatebox[origin=c]{0}{
                    \shortstack{EXPTM}}}
            &85.12 &\textbf{-1.40}&  85.99 &\textbf{-0.25}& 85.55 &\textbf{-0.84}\\
             
              \multirow{1}{*}{
                \rotatebox[origin=c]{0}{
                    \shortstack{JNLPBA}}}
            &75.85 &{+0.19}&  82.37 &\textbf{-0.74}& 78.98 &\textbf{-0.25}\\
             
             \multirow{1}{*}{
                \rotatebox[origin=c]{0}{
                    \shortstack{LINNAEUS}}}
            &88.16 &{+0.31}&  87.91 &{+0.56}& 88.03 &{+0.44}\\
             
             \multirow{1}{*}{
                \rotatebox[origin=c]{0}{
                    \shortstack{NCBI-DISEASE}}}
            &88.00 &\textbf{-1.34}&  90.46 &{+0.42}& 89.22 &\textbf{-0.50}\\
             
             \multirow{1}{*}{
                \rotatebox[origin=c]{0}{
                    \shortstack{2010-i2b2}}}
            &94.96 &{+0.31}&  95.71 &{+0.20}& 95.33 &{+0.26}\\

             \multirow{1}{*}{
                \rotatebox[origin=c]{0}{
                    \shortstack{2011-i2b2}}}
            &94.01 &{+0.41}&  94.09 &{+0.28}& 94.05 &{+0.35}\\

            \multirow{1}{*}{
                \rotatebox[origin=c]{0}{
                    \shortstack{2012-i2b2}}}
            &82.97 &\textbf{-1.64}&  86.65 &\textbf{-2.13}& 84.77 &\textbf{-1.88}\\
             \hline

            \end{tabular}
            }
            \caption{Precision (P), Recall (R) and F-Measure (F) using BERT-CNN model trained on \textbf{balanced  dataset} with \textbf{Question} as knowledge. $\Delta$P, $\Delta$R, $\Delta$F represent change in performance when compared to training our model on full datasets. Negative value indicates training on unbalanced dataset is better  while positive value indicates balanced dataset training produces better performance.  Negative  values are in bold.
                }
            \label{tab:sup_balanced}
        \end{table}

\begin{table*}
        \centering
        \small
        \scriptsize
        \resizebox{2\columnwidth} {!}{
        \begin{tabular}{l l r r r r r r r}
            \toprule 
            \textbf{Dataset} & \multicolumn{1}{c}{\textbf{Entity}} & \multicolumn{1}{c}{\textbf{Entity Mentions}} & \multicolumn{1}{c}{\textbf{Train + }} & \multicolumn{1}{c}{\textbf{Train - }} & \multicolumn{1}{c}{\textbf{Dev +}} & \multicolumn{1}{c}{\textbf{Dev -}}  & \multicolumn{1}{c}{\textbf{Test +}} & \multicolumn{1}{c}{\textbf{Test -}}\\
            \hline
            
             \multirow{1}{*}{\rotatebox[origin=c]{0}{\textbf{\texttt{AnatEM}}}}
            & A\textsc{natomy}  &13701 &3514&2169 & 1122 & 959 & 2308 & 1405 \\
            \hline
            
             \multirow{1}{*}{\rotatebox[origin=c]{0}{\textbf{\texttt{BC2GM}}}}
            & G\textsc{ene/}P\textsc{rotein}  &24516& 6404 & 6071 & 1283 & 1214 & 2568 & 2424 \\
            \hline
            
             \multirow{1}{*}{\rotatebox[origin=c]{0}{\textbf{\texttt{BC4CHEMD}}}}
            & C\textsc{hemical}  & 84249& 14488 & 16002 & 14554 & 15909 & 12415 & 13738 \\
            \hline
            
             \multirow{2}{*}{\rotatebox[origin=c]{0}{\textbf{\texttt{BC5CDR}}}}
            & C\textsc{hemical}  &14913 &2951 & 1595 & 3017 & 1551 & 3090 & 1688 \\
            & D\textsc{isease}  & 12852 &2658 & 1888 & 2727 & 1841 & 2842 & 1936 \\
            \hline
            
             \multirow{1}{*}{\rotatebox[origin=c]{0}{\textbf{\texttt{BioNLP09}}}}
            &G\textsc{ene/}P\textsc{rotein} &14963 & 4711&2716 & 1014 & 433 & 1700 & 739\\
            \hline
            
             \multirow{1}{*}{\rotatebox[origin=c]{0}{\textbf{\texttt{BioNLP11EPI}}}}
            &G\textsc{ene/}P\textsc{rotein} & 15881& 3797 & 1896 & 1241 & 714 & 2836 & 1282\\
            \hline
            
            \multirow{4}{*}{
                \rotatebox[origin=c]{0}{
                    \shortstack{\textbf{\texttt{BioNLP11ID}}}}}
            & G\textsc{ene/}P\textsc{rotein} & 6551&1255&1193 & 446& 265& 955 & 977\\ 
            & O\textsc{rganism}   &3469& 1120 & 1328 & 270& 441& 779 & 1153 \\
            & C\textsc{hemical}  &973 & 334 & 2114 & 77 & 634 & 151 & 1781\\
            & R\textsc{egulon-}O\textsc{peron}  &87&9& 2439 & 19&692 & 43 & 1889 \\
             \hline

            \multirow{16}{*}{
                \rotatebox[origin=c]{0}{
                    \shortstack{\textbf{\texttt{BioNLP13CG}}}}}
            & G\textsc{ene/}P\textsc{rotein} &7908 & 1956 & 1077 & 393 & 610 & 1185 & 721 \\ 
            & C\textsc{ell}   &4061& 1388 & 1645 & 399 & 604&714 & 1192  \\
            & C\textsc{hemical}  &2270 & 645 &2388 & 274& 729 & 431 & 1475 \\
            & C\textsc{ancer}  &2582& 908& 2125 &324 & 679 & 665 & 1241\\
            & O\textsc{rgan}  &2517& 919 & 2114 & 305 & 698 & 565 & 1341\\
            & O\textsc{rganism}   &2093& 827 & 2206 & 267 & 736 & 486 & 1420\\
            & T\textsc{issue}  &587 &259 & 2774&77 &926& 153 & 1753\\
            & A\textsc{mino} A\textsc{cid}  &135&38 & 2995 & 17 & 986 & 34 & 1872\\
            & C\textsc{ellular} C\textsc{omponent}  &569& 247 &2786 & 78 & 925 &138 & 1768\\
            & O\textsc{rganism} S\textsc{ubstance} &283& 131 & 2902&33 & 970  & 81 & 1825\\
            & P\textsc{athological} F\textsc{ormation}   &228& 91 & 2952& 35 & 968 & 73 & 1833 \\
            & A\textsc{natomical} S\textsc{ystem}  &41 & 16 & 3017 & 3  & 1000 & 17 & 1889\\
            & I\textsc{mmaterial} A\textsc{natomical} & 102& 47 & 2986 & 18 & 985  & 29 & 1877\\ 
            & O\textsc{rganism} S\textsc{ubdivision}   &98& 42 & 2991 &  12 & 991 & 35 & 1871\\
            & M\textsc{ulti-}T\textsc{issue} S\textsc{tructure} &857& 345 & 2688 & 114 & 889 &236 & 1670\\ 
            & D\textsc{eveloping} A\textsc{natomical} S\textsc{tructure}  & 35&13 & 3020 & 5  & 998 & 17 & 1889\\
             \hline
             
             \multirow{1}{*}{\rotatebox[origin=c]{0}{\textbf{\texttt{BioNLP13GE}}}}
            & G\textsc{ene/}P\textsc{rotein}  &12031&1499&901 & 1655 & 1010 & 1936 & 1376 \\
             \hline

             \multirow{4}{*}{\rotatebox[origin=c]{0}{\textbf{\texttt{BioNLP13PC}}}}
            & G\textsc{ene/}P\textsc{rotein}  & 10891& 2153& 346 & 723 & 134 & 1396 & 298 \\
            & C\textsc{omplex}  & 1502& 542 & 1957 & 178 & 679 & 398 & 1296 \\
            & C\textsc{hemical}  & 2487& 596 & 1903 & 244 & 613 & 450 & 1244\\
            & C\textsc{ellular/} C\textsc{omponent} &1013 & 373 & 2126 & 144 & 713 & 263 & 1431\\
            \hline

             \multirow{6}{*}{\rotatebox[origin=c]{0}{\textbf{\texttt{CRAFT}}}}
            & G\textsc{ene/}P\textsc{rotein}  & 16108& 4458 & 5539 & 1358 & 2105 & 3140 & 3634 \\
            & T\textsc{axonomy}  &6835 &2511 &7486 & 994 & 2469 & 1710 & 5064 \\
            & C\textsc{hemical}  & 6018& 1908 & 8089 & 586 & 2877 & 1344 & 5430 \\
            & C\textsc{ell} L\textsc{ine}  & 5487& 2058& 7939 & 540 & 2923 & 1257 & 5517\\
            & S\textsc{equence} O\textsc{ntology}  &18856 & 4303 & 5694 & 1711 & 1752 & 3023 & 3751 \\
            & G\textsc{ene} O\textsc{ntology}  & 4166& 1499 & 8498 & 336 & 3127 & 1344 & 5430 \\
             \hline

             \multirow{1}{*}{\rotatebox[origin=c]{0}{\textbf{\texttt{EXPTM}}}}
            & G\textsc{ene/}P\textsc{rotein} &4698& 857 & 520 & 279 & 158 & 1160 & 679 \\
            \hline
             
            \multirow{5}{*}{
                \rotatebox[origin=c]{0}{
                    \shortstack{\textbf{\texttt{JNLPBA}}}}}
            & D\textsc{NA}  & 10550& 4670 & 12146 & 553 & 1218 & 624 & 3226\\ 
            & R\textsc{NA}   &1061&713 & 16103 & 89& 1682 & 102 & 3748 \\
            & C\textsc{ell} L\textsc{ine} & 4315&2591&14225 & 285 & 1486 & 378 & 3472\\
            & C\textsc{ell} T\textsc{ype}  &8584& 4735 & 12081 & 415 & 1356 & 1403 & 2447\\
            & G\textsc{ene/}P\textsc{rotein}  &35234& 11840 & 4976 & 1137 & 634 & 2368 & 1482\\
             \hline

             \multirow{1}{*}{\rotatebox[origin=c]{0}{\textbf{\texttt{Linnaeus}}}}
            & S\textsc{pecies}  &4242 &1546 & 9173 & 520 & 3300 & 1029 &5381\\
            \hline
            
             \multirow{1}{*}{\rotatebox[origin=c]{0}{\textbf{\texttt{NCBI-Disease}}}}
            & D\textsc{isease}  &6871 &2921&2473 & 489 & 434 & 538 & 398 \\
             \hline

             
             \multirow{3}{*}{\rotatebox[origin=c]{0}{\textbf{\texttt{2010-i2b2}}}}
            & P\textsc{roblem}  &18979 & 4213 & 4226 & - &  -  & 5802 & 6590 \\
            & T\textsc{reatment}  & 13809&3126 & 4226& - & - & 7234 & 6590 \\
            & T\textsc{est}  &13576 & 2426 &4226& - & - &4591 & 6590 \\
            \hline
            
             \multirow{4}{*}{\rotatebox[origin=c]{0}{\textbf{\texttt{2011-i2b2}}}}
            & P\textsc{erson}  &17744 & 7207 & 3990& - &- & 4715 & 2971 \\
            & P\textsc{roblem}  & 18869 &7003 & 3990& - &- & 4384 &2971\\
            & T\textsc{reatment}  &17708 &5300  & 3990 & - &- & 3565 &  2971\\
            & T\textsc{est}  &13514 &4191&3990  & - &- &  2786  &2971\\
            \hline
             
             \multirow{6}{*}{\rotatebox[origin=c]{0}{\textbf{\texttt{2012-i2b2}}}}
            & P\textsc{roblem}  &4754& 2832 &3597& - & - & 2326 & 2683 \\
            & T\textsc{reatment}  & 7076&2341&4088& - & - & 1976 & 3033  \\
            & T\textsc{est}  & 4754& 1786&4643&  - & - & 1465 & 3544 \\
            & O\textsc{ccurance}  & 5126&2086&4343&  - & - & 1677 & 3332 \\
            & C\textsc{linical}-D\textsc{epartment} E\textsc{vent}  & 1716 &852& 5577 & - &-& 655& 4354  \\
            & E\textsc{vidential} E\textsc{vent}  &1334 &706&5723& - & - & 560 & 4449  \\
            \hline

            \end{tabular}
            }
            \caption{
            Data Distribution, with counts of entities, number of positive samples with at least one entity mentions, and negative samples with no target entity. 
                }
            \label{tab:data}
        \end{table*}



\end{document}